\documentclass[]{article}
\usepackage[a4paper, total={6in, 8in}]{geometry}
\usepackage{amssymb}
\usepackage{times}
\usepackage{epsfig}
\usepackage{graphicx}
\usepackage{amsmath}
\usepackage{times}
\usepackage{epsfig}
\usepackage{graphicx}
\usepackage{amsmath}
\usepackage{bbm}
\usepackage{algorithm}
\usepackage[noend]{algpseudocode}

\DeclareMathOperator*{\argmax}{argmax} 
\usepackage{multirow}
\usepackage{bigstrut}
\usepackage{xcolor}
\usepackage{booktabs}
\usepackage{subfig}
\usepackage{colortbl}
\usepackage{enumerate}
\usepackage{sidecap}
\usepackage{mathtools}
\usepackage{color}
\definecolor{highlight}{rgb}{0,0,0}
\usepackage{marvosym}
\title{Towards a category-extended object detector with limited data}

\author{Bowen Zhao, Chen Chen, Xi Xiao, Shu-Tao Xia}
\date{}
\usepackage{authblk}

\author[a]{Bowen Zhao}
\author[b,\Letter]{Chen Chen}
\author[a]{Xi Xiao}
\author[a,c,\Letter]{Shu-Tao Xia}
\affil[a]{Tsinghua Shenzhen International Graduate School, Tsinghua University}
\affil[b]{TEG AI, Tencent}
\affil[c]{Research Center of Artificial Intelligence, Peng Cheng Laboratory}
\affil[ ]{
\textit{zbw18@mails.tsinghua.edu.cn} \\
\textit{beckhamchen@tencent.com}\\
{\{xiaox,xiast\}@sz.tsinghua.edu.cn}
}
\begin{document}
\maketitle
% \begin{frontmatter}
\let\thefootnote\relax\footnotetext{\Letter \ Corresponding authors.}


% \affiliation[label1]{organization={},
%             city={Shenzhen},
%             country={China}}

% \affiliation[label2]{organization={},
%             city={Shenzhen},
%             country={China}}
            
% \affiliation[label3]{organization={},
%             city={Shenzhen},
%             country={China}}

\begin{abstract}
Object detectors are typically learned on fully-annotated training data with fixed predefined categories. However, categories are often required to be increased progressively. Usually, only the original training set annotated with old classes and some new training data labeled with new classes are available in such scenarios. Based on the limited datasets, a unified detector that can handle all categories is strongly needed. We propose a practical scheme to achieve it in this work. A conflict-free loss is designed to avoid label ambiguity, leading to an acceptable detector in one training round. To further improve performance, we propose a retraining phase in which Monte Carlo Dropout is employed to calculate the localization confidence to mine more accurate bounding boxes, and an overlap-weighted method is proposed for making better use of pseudo annotations during retraining. Extensive experiments demonstrate the effectiveness of our method.
\end{abstract}



% \end{frontmatter}

% \linenumbers


\section{Introduction}

\begin{figure}[t]
   \centering
   \includegraphics[width=0.55\textwidth]{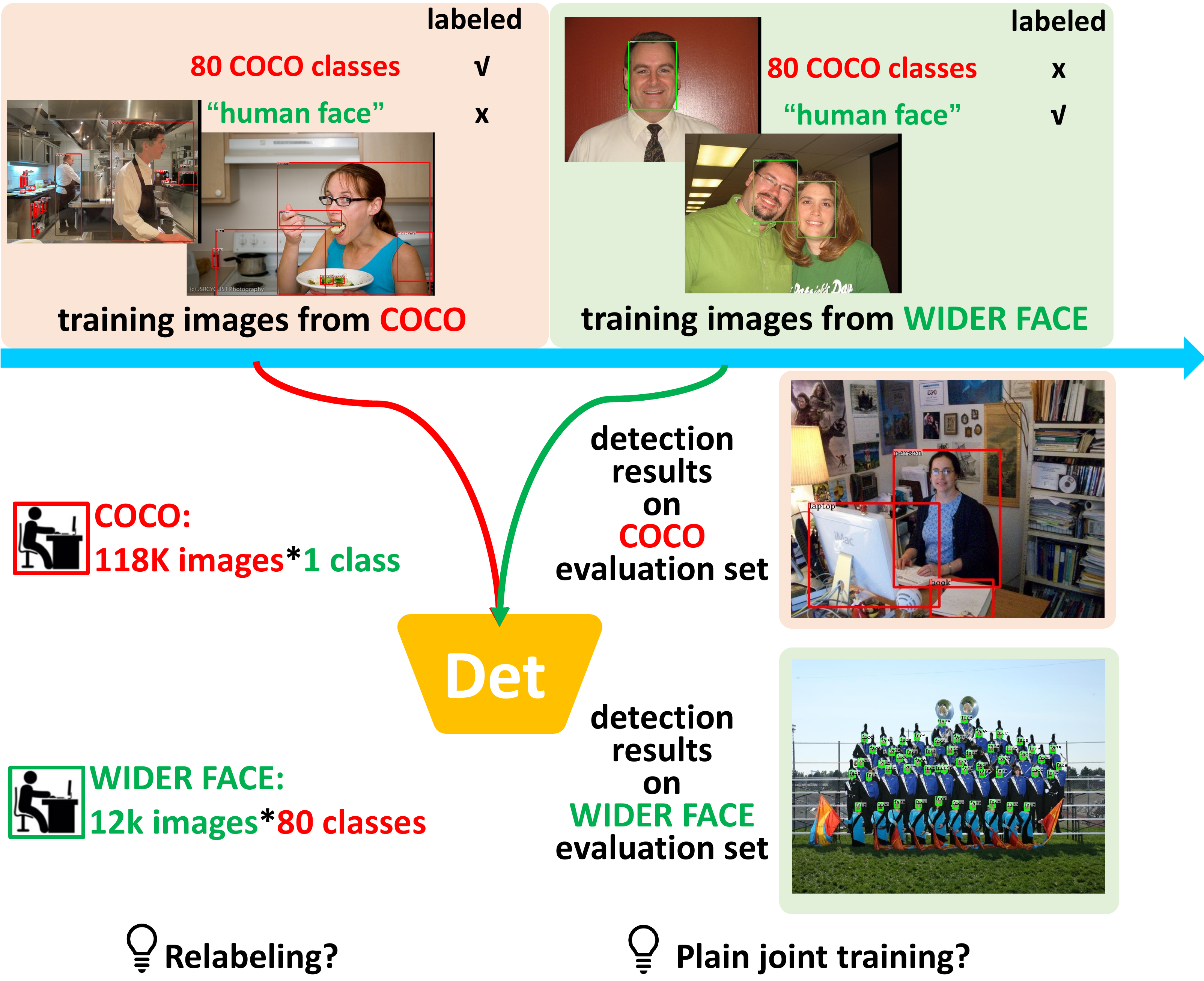}   \caption{A unified detector is expected to detect all categories (e.g., both the classes in {\color{red} COCO} and {\color{green} WIDER FACE}) while only separately labeled training sets are provided (``faces" are not labeled in COCO; simultaneously, instances of the 80 COCO classes are not annotated in FACE). Typically, two simple ways can achieve this goal: relabeling or plain joint training. However, relabeling is extremely costly (which needs to browse each image to discover instances belonging to the missing categories) and plain joint training leads to severe biases and a poor detector (which fails to detect ``face" in COCO evaluation set and ``person" in FACE evaluation set). Best seen in color.}
   \label{fig:introduction}
\end{figure}

Object detection is a fundamental and essential computer vision problem in various application areas, such as self-driving~\cite{WEI2020107195}, medicine~\cite{wang2020focalmix} and security~\cite{AKCAY2022108245}, etc. Owing to the development of deep neural networks, more and more powerful detectors are proposed~\cite{ren2015faster,lin2017focal,QIN2020107404}. 

A standard object detector is typically trained on a pre-prepared fully annotated training dataset with fixed categories. However, in many real-world applications, we cannot know all possible categories of interest beforehand, so it is often desired to add new object classes progressively. For instance, we have a detector in use, which was trained on an original dataset (e.g., COCO~\cite{lin2014microsoft} with 80 classes). Now, it is required to detect a new class ``human face" (abbr.: face) simultaneously while only some new training data labeled with ``face'' is additionally provided (e.g., WIDER FACE~\cite{yang2016wider}). As shown in Fig.~\ref{fig:introduction}, many faces in the original training data are not labeled, similarly, the instances belonging to the original categories (e.g., ``person") are not annotated in the new dataset. 

As new categories are incrementally added, training data is collected and labeled specifically for them. {\color{highlight}Commonly, the models are expected to be efficient, namely smaller, faster and better~\cite{redmon2016you,yang2020distilling,yang2020factorizable,jing2020dynamic,jing2021amalgamating}}. An important question is how to train a unified detector that can manage all categories based on the limited datasets, because a unified detector introduces merely negligible extra computation and memory overheads with respect to the number of categories of interest. As shown in Fig.~\ref{fig:introduction}, there are two straightforward ways to achieve a unified detector. The one is relabeling, which transforms the problem into a standard object detection task. However, it's prohibitively expensive to relabel all missing instances in the original and new datasets. For massive amounts of data and ever-increasing categories of interest, browsing images from all datasets to discover instances belonging to the missing categories is impractical. The other one is plain joint training. While, in one dataset (e.g., COCO), samples related to unlabeled instances (e.g., ``face") will be regarded as negative (background) samples, resulting in conflicts and false gradients during training. Due to overfitting the false negative samples, it is difficult for the detector trained in the plain way to discover categories not labeled in corresponding training data. For example, {\color{highlight} as shown in Fig.~\ref{fig:introduction}}, the model trained on the combined training set of COCO and WIDER FACE cannot detect ``face" in the COCO evaluation set or ``person" in the WIDER FACE evaluation set.

Can we obtain a unified detector that can detect all categories well while avoiding relabeling manually and alleviating conflicts during training? We propose a general solution for this purpose in the work. (i) A conflict-free loss is carefully designed to avoid two possible conflicts during joint learning (details in Sec.~\ref{sec:cf_loss}): (a) Samples assigned as negative in one dataset may be actually positive instances belonging to classes from the other datasets. (b) Samples assigned as positive may be more suitable to be labeled as categories of the other datasets. Experimental results demonstrate that conflict-free loss can lead to an acceptable unified detector. (ii) To further improve performance, we propose a retraining phase (details in Sec.~\ref{sec:gt_mining} and Sec.~\ref{sec:retraining}). (a) We attempt to mine high-quality unlabeled ground-truth with the trained detector. The Monte Carlo Dropout is employed to obtain localization confidence, which is combined with the classification confidence to unearth pseudo annotations with more accurate bounding boxes. (b) Regarding the retraining strategies with the mined pseudo annotations, we realize that although a lot of unlabeled instances are mined, there still hide many latent ground-truth. Therefore, the conflict-free idea cannot be removed in the retraining stage. However, simply copying the training strategy from the first phase results in insufficient negative information, as only positive samples are added by pseudo annotations during retraining. Consequently, we propose an overlap-based negative samples weighting strategy to utilize negative samples modestly in the retraining process. In the end, we achieve a strong and unified detector.

{\color{highlight} Object detection, semi-supervised object detection, open-set object detection, and class incremental object detection are all related to the study of category-extended object detector. A significant amount of research has been conducted in these areas. Sec.~\ref{sec:related_work} provides an abridged overview of the relevant background.}

To sum up, the main contributions of this paper are listed as follows.
\textbf{(i)} We present a general solution for training the unified detector with only the original datasets and incrementally labeled new datasets, which is urgently required in realistic applications.
\textbf{(ii)} A conflict-free loss is proposed to avoid possible label ambiguity. Localization confidence is designed for mining more accurate pseudo annotations. An overlap-weighted loss is employed to deal with uncertain negative samples for retraining a stronger detector with pseudo annotations. 
\textbf{(iii)} The entire pipeline does not introduce any additional manual labeling or modification of network structures and does not alter inference speed.
\textbf{(iv)} Extensive experiments are conducted to demonstrate the effectiveness of our proposed solution as compared with state-of-the-art approaches.

\section{Related Work}
\label{sec:related_work}

{\color{highlight} \paragraph{Object Detection} Object detection has made significant progress in the past decade. Detectors (R-CNN serials~\cite{ren2015faster,girshick2014rich,girshick2015fast}, RetinaNet~\cite{lin2017focal}, FCOS~\cite{tian2019fcos}, etc.) mainly focus on efficiency or performance enhancement based on sufficient and complete training data. Recently, object detection in more complex and realistic scenarios has also attracted much attention~\cite{wang2018geometry,yang2021training,gao2022discrepant}}.

\paragraph{Semi-Supervised Object Detection} SSOD aims to learn detectors based on a few labeled images and large amounts of unlabeled images. CSD~\cite{jeong2019consistency} and FocalMix~\cite{wang2020focalmix} utilize consistency regularization to make full use of the unlabeled data. Self-training and strong data augmentations are employed in STAC~\cite{sohn2020simple} to enhance the detector for SSOD. Although the issue of incomplete training data is also involved in this work, the main difference is that no extra images are introduced during training except the available partially-labeled datasets in our problem.

\paragraph{Open-Set Object Detection} OSOD~\cite{hall2020probabilistic,miller2018dropout} refers to the issue that objects from a new domain that are not seen in the training data may be encountered in the test data, which bring about more false positive predictions compared to the closed-set. In this work, there probably exists a non-ignorable domain gap between the new annotated training set and the original training set. Methods for OSOD inspire us to mine higher-quality pseudo annotations with fewer false positive predictions based on uncertainty estimation.

{\color{highlight}\paragraph{Class Incremental Object Detection} In recent years, class incremental learning has attracted much attention in diverse tasks, such as classification~\cite{rebuffi2017icarl,wu2019large,zhao2020maintaining,zhao2022energy}, segmentation~\cite{michieli2021knowledge} and also object detection. CIOD aims at developing a lifelong-learning detection system. In~\cite{shmelkov2017incremental,hao2019end,perez2020incremental,kj2021incremental}, efforts are made to learn detectors incrementally based on only new training images with new classes, i.e., the original training data is not available when new classes arrive. Specifically, knowledge distillation is employed to prevent the forgetting phenomenon in~\cite{shmelkov2017incremental} and the method in~\cite{hao2019end} further extend it to an end-to-end training manner. A meta-learned gradient preconditioning is proposed in~\cite{kj2021incremental} to not only minimize forgetting but also maximize knowledge transfer. In CIOD, the primary challenge is the catastrophic forgetting~\cite{mccloskey1989catastrophic,lao2021focl} for old classes due to a lack of original datasets. While the original data is still available in this work, we focus on a more immediate and realistic situation. Although it relaxes the constraints, we show that it is still a challenging problem (as shown in Fig.~\ref{fig:introduction}) which is frequently encountered in real-world applications and must be addressed urgently.}

\paragraph{Multi-Dataset Object Detection} MDOD tries to train a single detector on multiple datasets, which is the same purpose as in this work. In~\cite{zhao2020object} and~\cite{rame2018omnia}, a pseudo labeling approach is exploited. Dataset-aware focal loss is proposed in~\cite{yao2020cross} for the multi-dataset training. Compared to these methods, we present a more powerful pipeline to deal with this problem, which concentrates on three main questions: how to avoid label conflicts effectively and comprehensively (answered in Sec.~\ref{sec:cf_loss}), how to mine more accurate pseudo annotations (answered in Sec.~\ref{sec:gt_mining}), and how to make better use of pseudo labels (answered in Sec.~\ref{sec:retraining}).

\section{Category-extended Object Detector}

In this section, we introduce our solution for training category-extended detectors. We analyze the possible label conflicts hidden in the plain joint training and propose a general loss formula (conflict-free loss) for this problem (Sec.~\ref{sec:cf_loss}). It attempts to take full advantage of the exact information and avoid ambiguous one, leading to an acceptable detector in one training round. Then, to further improve the performance, we design the retraining phase. It consists of two components: (a) Unlabeled ground-truth mining with classification and localization confidence (CLC) (Sec.~\ref{sec:gt_mining}). The CLC-based unlabeled ground-truth mining method helps to obtain more accurate pseudo annotations. (b) Retraining with overlap-weighted negative samples (Sec.~\ref{sec:retraining}). The overlap-weighted negative samples in retraining help to make full use of the pseudo annotations and obtain a better detector.

\subsection{Notations}
For brevity of presentation and without loss of generality, we assume that there are original dataset $\mathcal{D}_o$ (denoted by categories $\mathcal{C}_o$, images $\mathcal{I}_o$ and ground-truth annotations $\mathcal{G}_o$) and newly-added dataset $\mathcal{D}_n$ (denoted by categories $\mathcal{C}_n$, images $\mathcal{I}_n$ and ground-truth annotations $\mathcal{G}_n$) with different label spaces\footnote{This technique is also applicable to multiple datasets.} ($\mathcal{C}_o$ and $\mathcal{C}_n$ do not include the special category ``background"). We aim to train a unified object detector on $\mathcal{D}_o$ and $\mathcal{D}_n$. The overall loss function for training can be formulated as a weighted sum of the classification loss ($\mathcal{L}_{cls}$) and the localization loss ($\mathcal{L}_{loc}$):
\begin{equation}
\mathcal{L}(\{\mathbf{p}_i\}, \{\mathbf{t}_i\};
\{\mathbf{p}^*_i\}, \{\mathbf{t}^*_i\}) =
\mathcal{L}_{cls}(\{\mathbf{p}_i\}, \{\mathbf{p}^*_i\})
+
\mathcal{L}_{loc}(\{\mathbf{t}_i\}, \{\mathbf{t}^*_i\}),
\label{eq:overall_loss}
\end{equation}
where $i$ is the index of an anchor, $\mathbf{t}_i$ is a $4$-dimensional vector representing the parameterized coordinates of the predicted bounding box, and $\mathbf{t}_i^*$ is that of the ground-truth box associated with a positive anchor. $\mathbf{p}_i^*$ and $\mathbf{p}_i$ are the ground-truth label and the predicted probability vector of anchor $i$, respectively. The localization loss $\mathcal{L}_{loc}$ usually can be Smooth $L_1$ loss~\cite{he2016deep}. The classification loss $\mathcal{L}_{cls}$ can be a binary cross-entropy (BCE) based loss. It views the classification task as a series of independent binary classification tasks, which is formulated as
\begin{equation}
  \mathcal{L}_{cls} \coloneqq \mathcal{L}_{bce}(\{ \mathbf{p}_i \}, \{ \mathbf{p}^*_i \}) =
   \frac{1}{N} \sum_{i,c}
   l({p^c_i, p^*_i}^c)
   ,
\label{eq:bce_loss}
\end{equation}
where $
l({p^c_i, p^*_i}^c) = -[{p^*_i}^c log(p^c_i) +
(1 - {p^*_i}^c) log(1 - p^c_i)]
$. Under the BCE-based loss, $\mathbf{p}_i^*$ is a $|\mathcal{C}_o \cup \mathcal{C}_n|$-dimensional vector. If anchor $i$ matches with a ground-truth box in $\mathcal{G}_o$ or $\mathcal{G}_n$ of category $c$, $\mathbf{p}_i^*$ is denoted as a one-hot vector with only ${p_i^*}^c=1$. If it does not match any box in $\mathcal{G}_o \cup \mathcal{G}_n$, $\mathbf{p}_i^*$ will be set to $\mathbf{0}$.

As an example shown in Fig.~\ref{fig:introduction}, the detector trained with plain joint training on the combined dataset of COCO (dose not labeled ``face") and WIDER FACE (dose not labeled ``person") cannot detect ``face" in the COCO evaluation set or ``person" in the WIDER FACE evaluation set. The detector is biased due to the datasets and training method: in one dataset (e.g., COCO), samples related to unlabeled instances (e.g., “face") will be regarded as negative (background) samples, resulting in conflicts and false gradients during training. To relieve this problem, in the next subsection, we propose conflict-free loss, which attempts to make full use of the correct information and avoid using ambiguous information.

\subsection{Conflict-Free Loss}
\label{sec:cf_loss}

\begin{figure}[t]
  \centering
  \includegraphics[width=0.8\textwidth]{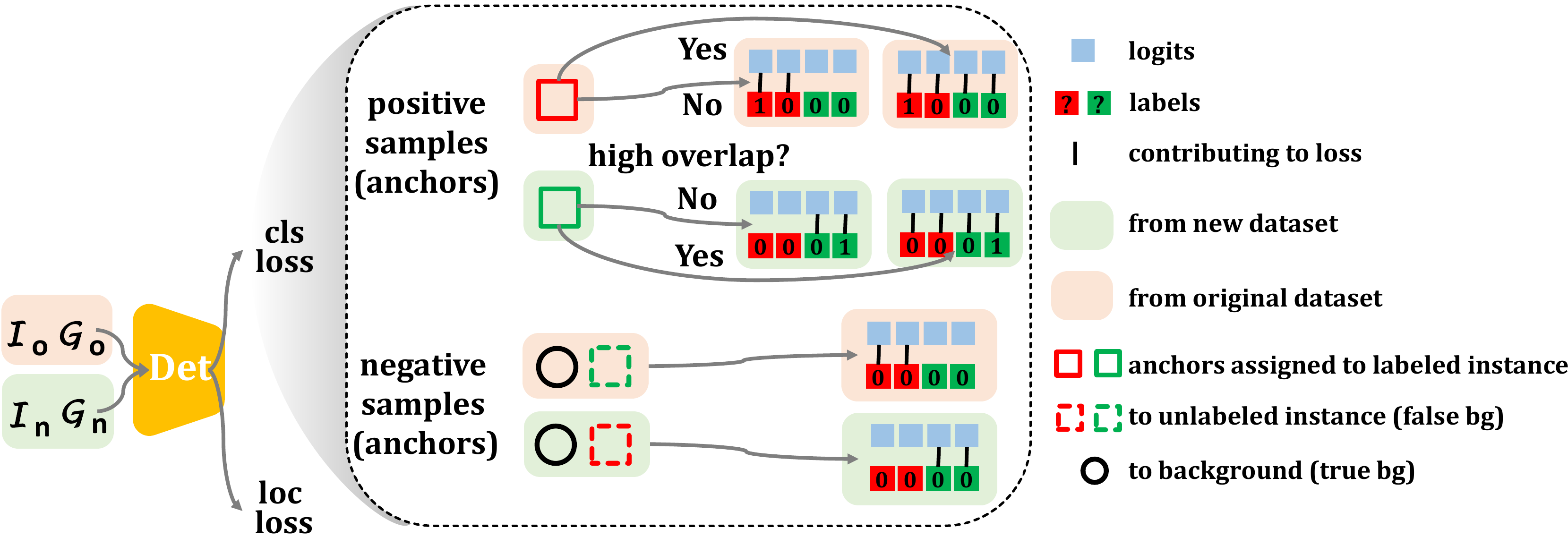}
  \caption{Illustration of conflict-free loss. Two possible conflicts are relieved, details in Sec.~\ref{sec:cf_loss}. {\color{red} Red} for original classes and {\color{green} green} for new classes. Best seen in color.}
  \label{fig:cf_loss}
\end{figure}

In Eq.~\eqref{eq:overall_loss}, the localization loss is activated only for positive anchors ($i \in \text{Pos}$) and disabled otherwise ($i \in \text{Neg}$), so that there are few possible conflicts hidden in it and we keep the localization loss unchanged. However, the plain BCE-based loss will be affected by false signals during joint training on multiple datasets. Thus, the classification loss should be carefully designed. There are two possible conflicts in plain BCE-based loss. (i) Anchors assigned as negative samples in one dataset may be positive samples belonging to unlabeled instances with classes from the other datasets. (ii) The assigned positive samples may belong to categories of other datasets. Commonly, in one dataset, an anchor is assigned as positive for one specific category when the Intersection over Union (IoU) with the ground-truth is greater than a threshold (say, 0.5 typically). However, such loose restriction may ignore the possibility that the anchor is more suitable to be labeled as another category that is only annotated in other datasets. 

Consequently, we propose the Conflict-Free (CF) loss, which is formulated as
\begin{equation}
  \mathcal{L}_{cls} \coloneqq \mathcal{L}_{cf}(\{ \mathbf{p}_i \}, \{ \mathbf{p}^*_i \}) =
  \frac{1}{N} \sum_{i,c} w(i, c) \cdot
  l({p^c_i, p^*_i}^c)
  ,
\label{eq:cf_loss}
\end{equation}
where, $w(i, c)$ is given by Alg.~\ref{algo:omega_training}, $\star$ represents $o$ or $n$, $f(i)$ represents the maximum IoU of anchor $i$ with ground-truth, $\tau_{s}$ is a strict threshold (0.9, conservatively). An illustration of Eq.~\eqref{eq:cf_loss} and Alg.~\ref{algo:omega_training} is shown in Fig.~\ref{fig:cf_loss}. In conflict-free loss, the two possible conflict origins are removed. First, a negative sample will not contribute to the loss of classes that are not labeled in its dataset $(i \in \mathcal{I}_{\star} \ \& \ c \notin \mathcal{C}_{\star} \ \& \ i \in \text{Neg})$. Second, the conflict-free loss does not concern the positive anchors with insufficient large overlaps with ground-truth ($\text{IoU} < \tau_{s}$) to provide negative information to classes from the other datasets $(i \in \mathcal{I}_{\star} \ \& \ c \notin \mathcal{C}_{\star} \ \& \ i \in \text{Pos} \ \& \ f(i) < \tau_{s})$. Note that the name ``conflict-free" is used to represent a general loss formula for avoiding conflicts, which can be used directly or integrated with any other BCE-based loss (e.g., Focal loss~\cite{lin2017focal}) by replacing $l$.

\begin{algorithm}[tp]
\centering
\caption{Obtaining $\{\omega (i,c)\}$ for Eq.~\eqref{eq:cf_loss}.}
\begin{algorithmic}[1]
\For{ anchor $i \in 1, \cdots, N $; class $c \in 1, \cdots, |\mathcal{C}_o \cup \mathcal{C}_n|$}
\If{$i \in \mathcal{I}_{\star} \ \& \ c \notin \mathcal{C}_{\star}$}
    \If{$i \in \text{Neg} \ | \ (i \in \text{Pos} \ \& \ f(i) < \tau_{s})$}
    \State $\omega(i,c)=0$
    \Else
    \State $\omega(i,c)=1$
    \EndIf
\Else
    \State $\omega(i,c)=1$
\EndIf
\EndFor
\State \textbf{Output: $\{\omega (i,c)\}$} 
\end{algorithmic}
\label{algo:omega_training}
\end{algorithm}

\begin{figure*}[t]
  \centering
  \includegraphics[width=0.65\textwidth]{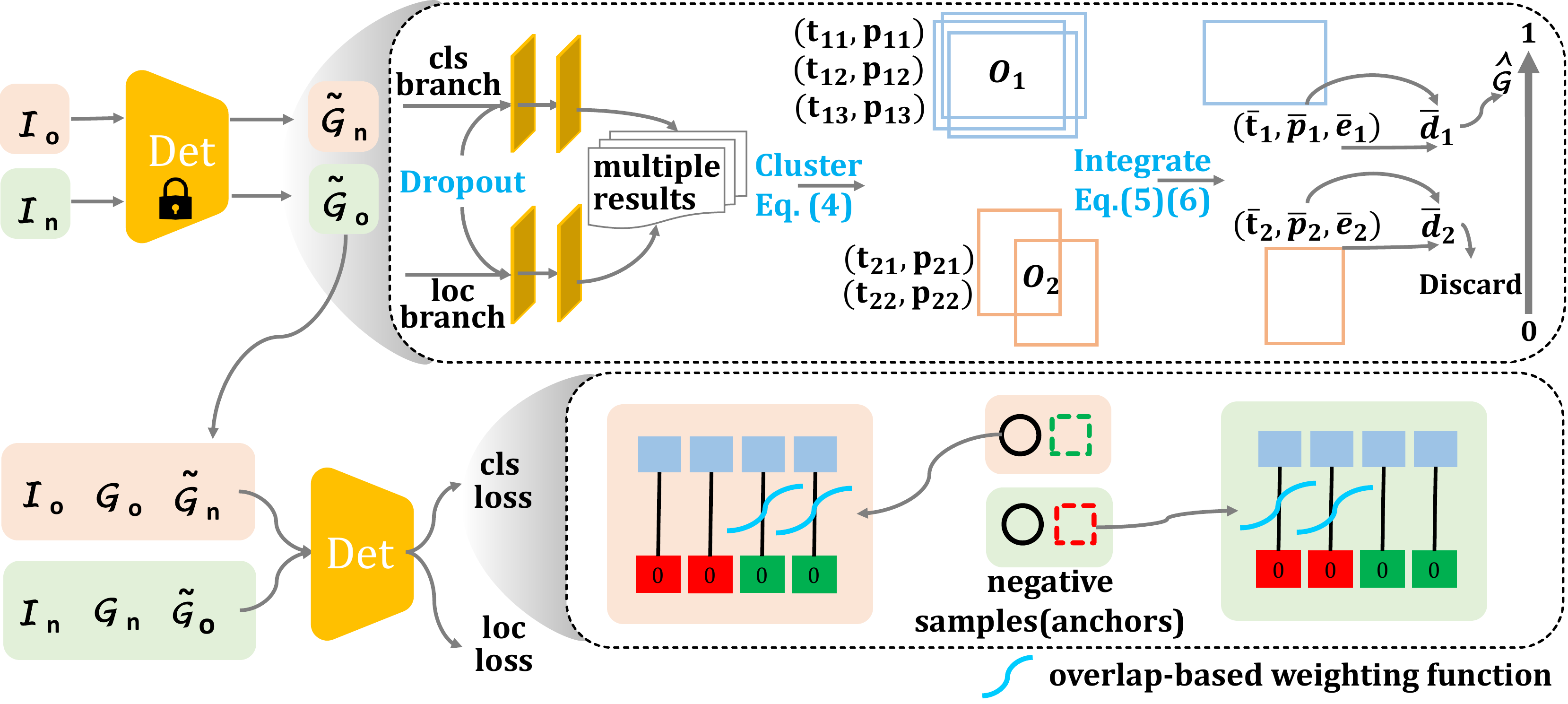}
  \caption{Illustration of the retraining method. (top) Unlabeled ground-truth mining with both classification and localization confidence. (bottom) Retraining on the supplemented datasets with overlap-based weighting schedule for negative samples. The remaining legend is the same as in Fig.~\ref{fig:cf_loss}. Best seen in color.}
  \label{fig:retraining}
\end{figure*}

It is demonstrated in the later experiments that an acceptable unified detector can be learned with conflict-free loss in one training round. While, there still exist many underutilized unlabeled objects. Inspired by self-training, a retraining phase is employed to further improve performance, which consists of two components: unlabeled ground-truth mining with classification and localization confidence (Sec.~\ref{sec:gt_mining}) and retraining with overlap-weighted negative samples (Sec.~\ref{sec:retraining}).

\subsection{Unlabeled Ground-Truth Mining}
\label{sec:gt_mining}

Unlike classification tasks, detectors need to predict the localization of objects, so that accurate localization annotations usually lead to better performance. Previous studies~\cite{jiang2018acquisition,he2019bounding} show that classification score cannot be interpreted as localization confidence: a predicted bounding box with high classification confidence may be localized inaccurately. If the inaccurate predicted results are adopted as pseudo annotations, they will prevent detectors from getting better performance. In pursuit of high-quality pseudo annotations, we propose the Classification and Localization Confidence (CLC) based unlabeled ground-truth mining method.

Monte Carlo Dropout-based Bayesian neural networks have shown great potential in measuring model uncertainty~\cite{gal2016dropout,kendall2017uncertainties,miller2018dropout,miller2019evaluating}. Inspired by this concept, during the inference phase, we perform $T$ forward passes of the base detector learned in Sec.~\ref{sec:cf_loss} with activated dropout layers as depicted in Fig.~\ref{fig:retraining}. Notice that since we introduce dropout operation only on the last layer of the classification and localization head while all the previous layers are identical and the computations are shared, the extra inference time during $T$ forward passes is ignorable. Then, for each image, we perform clustering on the bounding boxes of each class, resulting in a set of clusters. Each cluster $\mathcal{O}_i$ is made up of several bounding boxes with their respective classification score, defined as
\begin{equation}
   \begin{aligned}
   \mathcal{O}_i &= \{(\mathbf{t}_{ij}, \mathbf{p}_{ij}) \ | \ j=1,2,\cdots,|\mathcal{O}_i| \leq T \}, \\
   \ \text{s.t.} &\ \text{IoU}(\mathbf{t}_{ij_1}, \mathbf{t}_{ij_2}) \geq \tau_{nms}, \forall \ \mathbf{t}_{ij_1}, \mathbf{t}_{ij_2} \in \mathcal{O}_i,\\
   &\argmax_{r \in \mathcal{C}_o \cup \mathcal{C}_n} \ p^r_{ij_1} = \argmax_{r \in \mathcal{C}_o \cup \mathcal{C}_n} \ p^r_{ij_2}, \forall \ \mathbf{p}_{ij_1}, \mathbf{p}_{ij_2} \in \mathcal{O}_i,
   \end{aligned}
\end{equation}
where the threshold for clustering is the same as the NMS IoU threshold $\tau_{nms}$ (0.5 typically). Then the cluster $\mathcal{O}_i$ is represented by a triplet $(\overline{\mathbf{t}}_i, \overline{p}_i, \overline{e}_i)$, where the integrated bounding box $\overline{\mathbf{t}}_i$ and classification confidence $\overline{p}_i$ can be calculated by
\begin{equation}
   \overline{\mathbf{t}}_i = \frac{1}{|\mathcal{O}_i|} \sum_{j} \mathbf{t}_{ij}, \
   \overline{p}_i = \frac{1}{|\mathcal{O}_i|} \sum_{j} \max(p^1_{ij}, \cdots, p^{|\mathcal{C}_o \cup \mathcal{C}_n|}_{ij}),
\end{equation}
More importantly, the localization confidence is given by
\begin{equation}
\overline{e}_i =
\frac{ 1 + \mathbbm{1}_{|\mathcal{O}_i| \geq \frac{T}{2}} }{2|\mathcal{O}_i|^2}
\sum_{j_1, j_2} \text{IoU}(\mathbf{t}_{ij_1}, \mathbf{t}_{ij_2}).
\label{eq:loc_conf}
\end{equation}

Then, we obtain the detection confidence $\overline{d}_i = \overline{p}_i \times \overline{e}_i$, which considers both classification and localization confidence, and can better reflect the quality of predictions. Finally, we accept an integrated bounding box as pseudo ground-truth for class $c$ only if it satisfies: (i) class $c$ is not labeled in the dataset originally; (ii) $\overline{d}_i$ is larger than a pre-defined threshold $\eta$. Generally speaking, only the predictions with enough high detection confidence will be selected as pseudo annotations for the unlabeled classes. Fig.~\ref{fig:retraining} (top) illustrates the mining method.

\subsection{Retraining with Pseudo Annotations}
\label{sec:retraining}
After acquiring the pseudo annotations, we try to train a more powerful unified detector with the ground-truth and pseudo annotations together. So how to make better use of pseudo annotations? The most straightforward way is viewing the combined dataset supplemented with pseudo annotations as fully labeled~\cite{sohn2020simple}. Then the detector can be trained in a normal way with common classification losses (Eq.~\eqref{eq:bce_loss}). However, there may still exist considerable instances that are difficult to be mined. Hence, there is a risk that the inaccurate supervisory information may be introduced. Instead, one can retrain the detector with the conflict-free loss again (Eq.~\eqref{eq:cf_loss} and Alg.~\ref{algo:omega_training}). Compared to the first phase, the negative samples still do not contribute to the loss of classes from different datasets, only positive samples are supplemented by the pseudo annotations. So that the imbalance between positive samples and negative samples may be intensified in this way. A solution to alleviate the imbalance problem mentioned above is to introduce some ``safe" negative samples~\cite{zhao2020object}, which can contribute to the loss of classes from different datasets while ``unsafe" negative samples are still discarded.

\begin{algorithm}[t]
\centering
\caption{Obtaining $\{\omega (i,c)\}$ for Eq.~\eqref{eq:cf_loss} during retraining.}
\begin{algorithmic}[1]
\For{ anchor $i \in 1, \cdots, N $; class $c \in 1, \cdots, |\mathcal{C}_o \cup \mathcal{C}_n|$}
\If{$i \in \mathcal{I}_{\star} \ \& \ c \notin \mathcal{C}_{\star} \ \& \ i \in \text{Neg}$}
    \State $\omega(i,c)= Gom(f(i))$
\Else
    \State $\omega(i,c)=1$
\EndIf
\EndFor
\State \textbf{Output: $\{\omega (i,c)\}$} 
\end{algorithmic}
\label{algo:omega_retraining}
\end{algorithm}

\begin{figure}[t]
  \centering
  \includegraphics[width=0.25\textwidth]{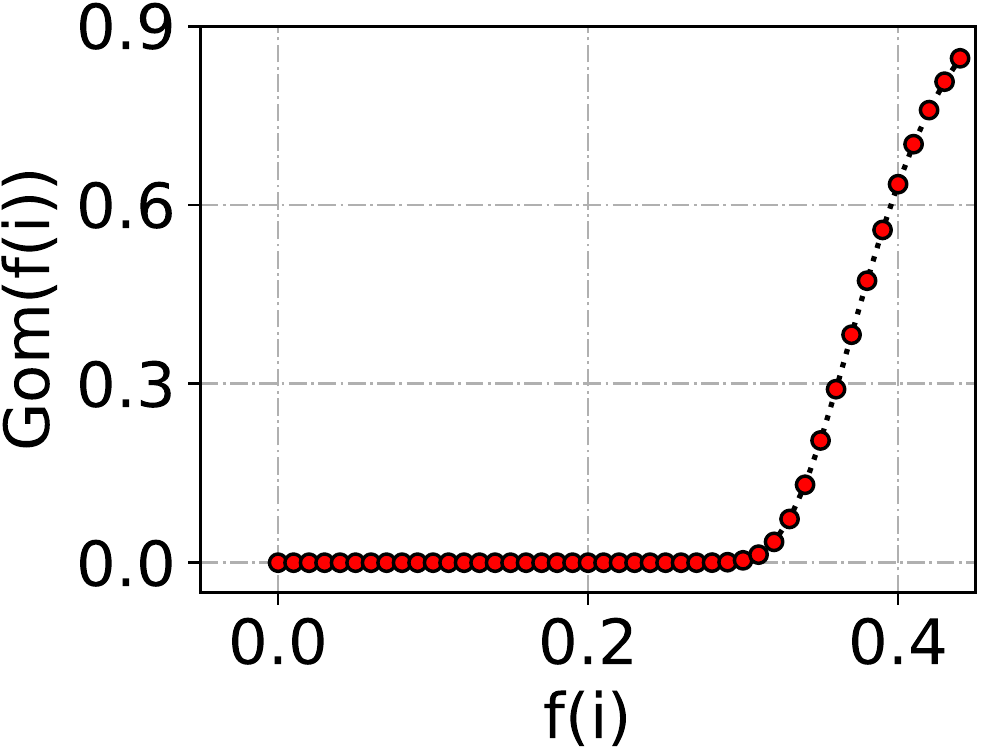}
  \caption{Overlap-based weighting schedule ($\tilde{a}=1$, $\tilde{b}=10,000$ and $\tilde{c}=25$). The negative samples' contribution to the loss of classes from different datasets is assigned with high weight only when they have a large overlap with annotated boxes.}
  \label{fig:gom_func}
\end{figure}

Instead, we introduce an overlap-based weighting schedule~\cite{wu2018soft} here to re-weight the negative samples' contribution to the loss of classes from different datasets. The main hypothesis is that negative samples which have relatively large overlaps (like 0.3 IoU, 0.4 IoU) with the existing ground-truth (or mined pseudo annotations) are probably not related to any unlabeled instances. On the contrary, negative samples with small overlaps have a higher risk of being unlabeled instances, i.e., false background. Based on this hypothesis, the Gompertz function
$$
Gom(x) = \tilde{a} e^{-\tilde{b} e^{-\tilde{c}x}}
$$
is employed to describe the relationship between overlap and the weight of loss, where $\tilde{a}$ is an asymptote, $\tilde{b}$ sets the displacement along the x-axis, $\tilde{c}$ sets the growth rate. We plot this function in Fig.~\ref{fig:gom_func} with $\tilde{a}=1$, $\tilde{b}=10,000$ and $\tilde{c}=25$. The negative samples' contribution to the loss of classes from different datasets is assigned with high weight only when they have a large overlap with annotated boxes. Other negative samples are still given very low weights. 

Our overlap-weighted retraining method can be formulated as training with Eq.~\eqref{eq:cf_loss} and $\{\omega (i,c)\}$ calculated from Alg.~\ref{algo:omega_retraining}. Considering many unlabeled instances are mined and the main concern in retraining is balancing the extra positive samples brought by pseudo annotations, we no longer consider the conflicts in positive samples in the retraining phase. In later experiments, we found that this method can take advantage of the mined pseudo annotations more fully.

\section{Experiments}

\subsection{Experimental Settings}
\label{sec:exp_set}

\paragraph{Datasets} We conduct experiments on several widely used datasets with different settings. MS COCO~\cite{lin2014microsoft} (2017) consists of 118,287 training images (COCO-train), 5,000 validation images (COCO-val) and 40,670 test images with 80 categories. PASCAL VOC~\cite{everingham2010pascal} (2007) has 9,963 images with 20 categories, 50\% for training/validation (VOC-train) and 50\% for testing (VOC-test). WIDER FACE~\cite{yang2016wider} is comprised of 32,203 images with one category ``face'', training 40\% (FACE-train), validation 10\% (FACE-val) and test 50\%. To evaluate the unified models, we also adopt the re-labeled evaluation sets \cite{zhao2020object}: VOC-subtest (a subset of VOC-test) contains 500 images with annotations for the 80 categories in COCO; FACE-subval (a subset of FACE-val) consists of 500 images with annotations for the ``face'' in FACE and the ``person'' in COCO; COCO-subval (a subset of COCO-val) has 500 images with annotations for the 80 categories in COCO and the ``face'' in FACE.

\begin{table}[t]
\centering
\caption{The statistics of the training sets used in our experiments.}
\resizebox{\textwidth}{!}{
\begin{tabular}{llllll}
\toprule
setup    &   training set    & \multicolumn{1}{l}{\#categories} & \multicolumn{1}{l}{\#images} & \multicolumn{1}{l}{\#annotations} & \multicolumn{1}{l}{\#missing annotations} \\
\midrule
A & COCO75-train & 75    & 100,543 & 601,410 & 129,089 \\
& COCO5-train & 5     & 17,744 & 22,976 & 106,526 \\
B & COCO79-train & 79    & 100,543 & 506,916 & 222,696 \\
& COCO1-train & 1     & 17,744 & 39,769 & 90,620 \\
C & COCO60-train & 60    & 118,287 & 367,189 & 492,812 \\
& VOC-train & 20    & 5,011 & 15,662 & unknown \\
D & COCO-train & 80    & 118,287 & 860,001 & unknown \\
& FACE-train & 1     & 12,880 & 159,424 & unknown \\
E & COCO60-train & 60    & 118,287 & 367,189 & unknown \\
& VOC-train & 20    & 5,011 & 15,662 & unknown \\
& FACE-train & 1     & 12,880 & 159,424 & unknown \\
\bottomrule
\end{tabular}%
}
\label{tab:data_stat}%
\end{table}%

\paragraph{Setups} To analyze the effect of our pipeline, we design five experimental setups. The statistics of the training sets used in these setups are summarized in Tab.~\ref{tab:data_stat}.

\textbf{Setup A}: COCO75-train and COCO5-train. We split COCO-train into two sets. One set is named COCO75-train, in which only annotations of 75 categories are retained, while annotations related to the other 5 categories are removed; another set, named COCO5-train, contains only annotations of the 5 categories. COCO75-train and COCO5-train are regarded as original dataset $\mathcal{D}_o$ and new dataset $\mathcal{D}_n$, respectively. In COCO75-train, there exist considerable unlabeled objects of the remaining 5 categories, similar in COCO5-train. We report the performance of the 75 classes and the 5 classes on COCO-val. Since there are more missing annotations belonging to the new 5 classes in the combined training set, we should pay more attention to the results of these 5 classes. 

\textbf{Setup B}: COCO79-train and COCO1-train. We perform another split on COCO-train: the original dataset COCO79-train contains 79 classes, and the new dataset COCO1-train has only one category. We report the performance of the 79 classes and the one class on COCO-val. Similarly, the results of the new class should be paid more attention.

\textbf{Setup C}: COCO60-train and VOC-train. In these experiments, we regard VOC-train as the new dataset and COCO60-train as the original dataset, which is generated by removing the annotations of the 20 VOC categories from COCO-train. We report the performance of all 80 classes on VOC-subtest and COCO-subval, and of the 20 VOC classes on COCO-val.

\textbf{Setup D}: COCO-train and FACE-train. We also train unified detectors on COCO-train (as original dataset) and FACE-train (as new dataset). The ``face'' is not annotated in COCO-train and the 80 COCO classes (``person'' mainly) are also not labeled in FACE-train. The performance of ``face" and ``person" on COCO-subval and FACE-subval is reported, in which the results of ``face" on COCO-subval and ``person" on FACE-subval should be paid more attention. 

\textbf{Setup E}, which involves three training sets labeled with different classes: COCO60-train, VOC-train, and FACE-train. The whole label space has 81 classes. We report the performance of all 81 classes on VOC-subtest and COCO-subval, and of ``person" on FACE-subval. 

\paragraph{Metrics} We use the standard metrics for object detection: AP, $\text{AP}_{50}$, and $\text{AP}_{75}$.

\paragraph{Implementations Details} 
For fair comparisons, all experiments are conducted on RetinaNet~\cite{lin2017focal} with ImageNet-pretrained ResNet-50~\cite{he2016deep} as the backbone. The models are implemented with PyTorch~\cite{paszke2019pytorch} and MMDetection~\cite{chen2019mmdetection}. In all experiments, input images are resized to $1333\times 800$. The models are trained using SGD over 8 GPUs with 2 images per GPU with 0.9 momentum, 0.0001 weight decay. We train 12 epochs in total with an initial learning rate of 0.02, and decrease the learning rate by 0.1 at epoch 8 and 11. For the conflict-free loss, the more strict threshold $\tau_{s}$ is set to 0.9 conservatively as described in Sec.~3.2 in all experimental settings. For the CC-based unlabeled ground-truth mining method, the confidence threshold $\eta$ is set to 0.5 in all experimental settings for best overall performance. For the CLC-based mining method, the number of forward passes $T$ is set to 20 for all experimental settings; the confidence threshold $\eta$ is set to 0.525 in the experiment on COCO75-train and COCO5-train, and 0.5 in other experimental settings for best overall performance. For the retraining strategy ``safe negatives", the lower confidence threshold $\eta'$ is set to 0.1 to obtain high-recall pseudo annotations in all experimental settings. For the retraining strategy ``overlap-weighted", we set $\tilde{a}=1$, $\tilde{b}=10,000$ and $\tilde{c}=25$ as shown in Sec.~3.4 in all experimental settings. There is almost no change in hyper-parameters in different experimental settings, which shows that our pipeline is robust.

\subsection{Overall Performance}

\begin{figure}[t]
  \centering
  \subfloat[]{\includegraphics[height=0.16\textwidth]{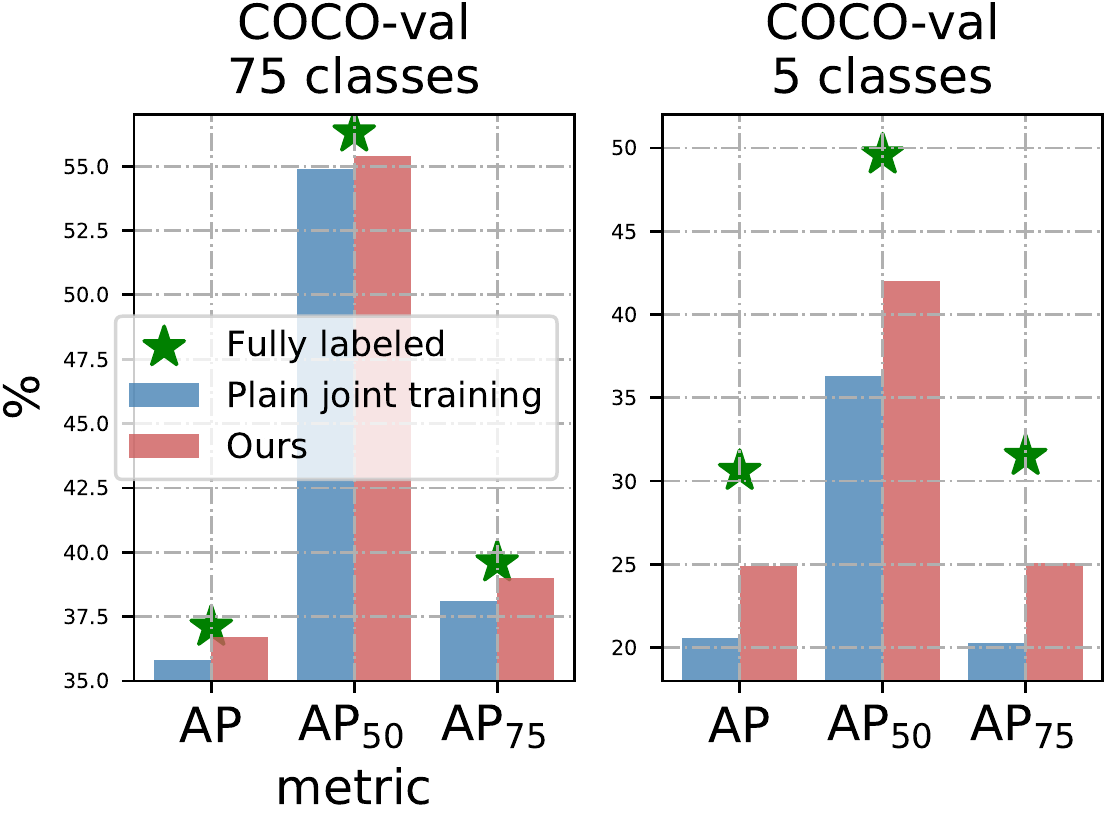}},
  \subfloat[]{\includegraphics[height=0.16\textwidth]{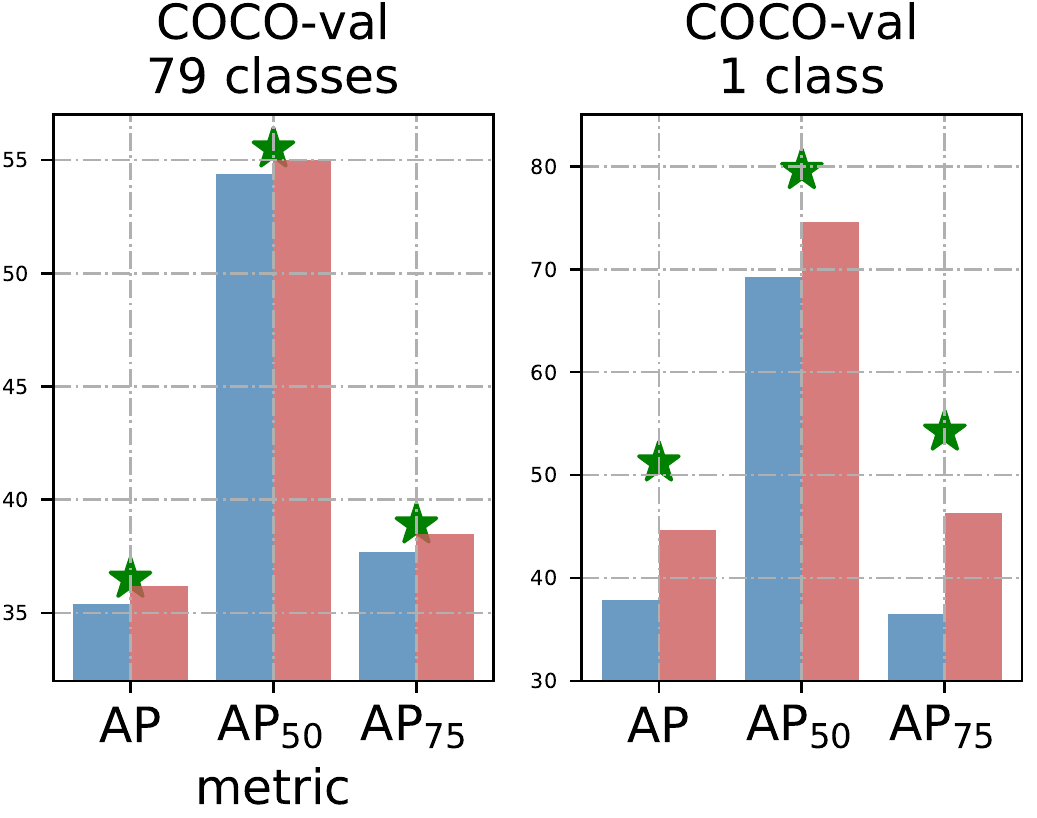}},
  \subfloat[]{\includegraphics[height=0.16\textwidth]{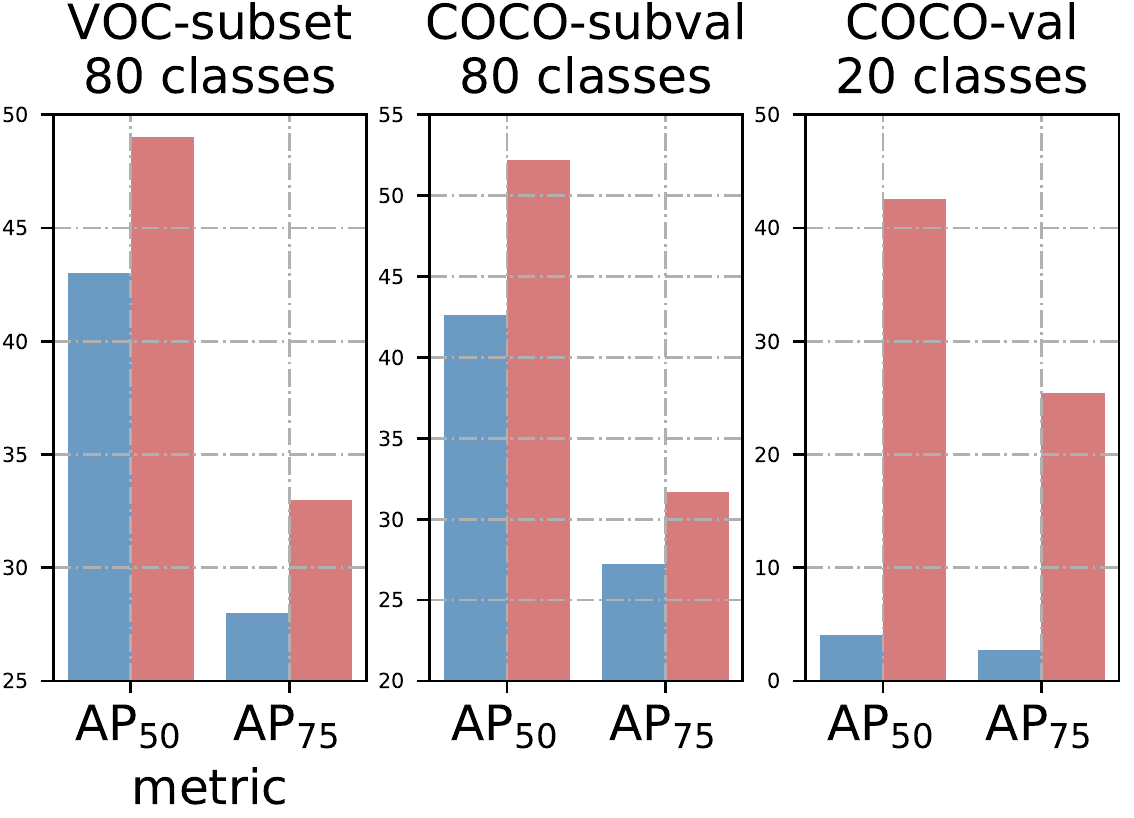}},
  \subfloat[]{\includegraphics[height=0.16\textwidth]{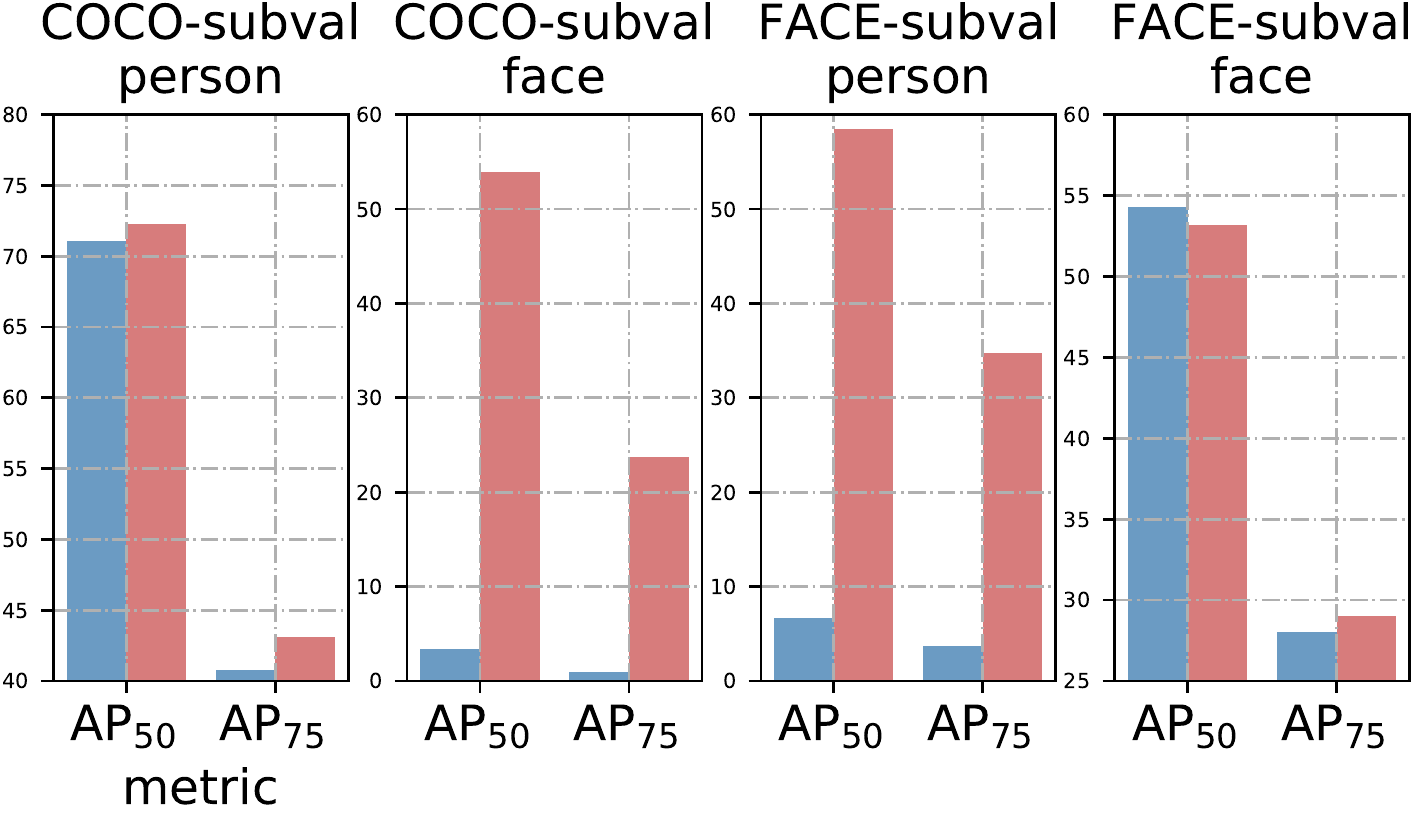}}
  \caption{Overall Performance. Subfigs (a), (b), (c) and (d) show the results of setup A, B, C and D, respectively. The performance is evaluated on specific sets and categories for each setup, details in Sec.~\ref{sec:exp_set}. Best seen in color.}
  \label{fig:compare_baselines}
\end{figure}

\begin{figure}[t]
  \centering
  \includegraphics[height=0.2\textheight]{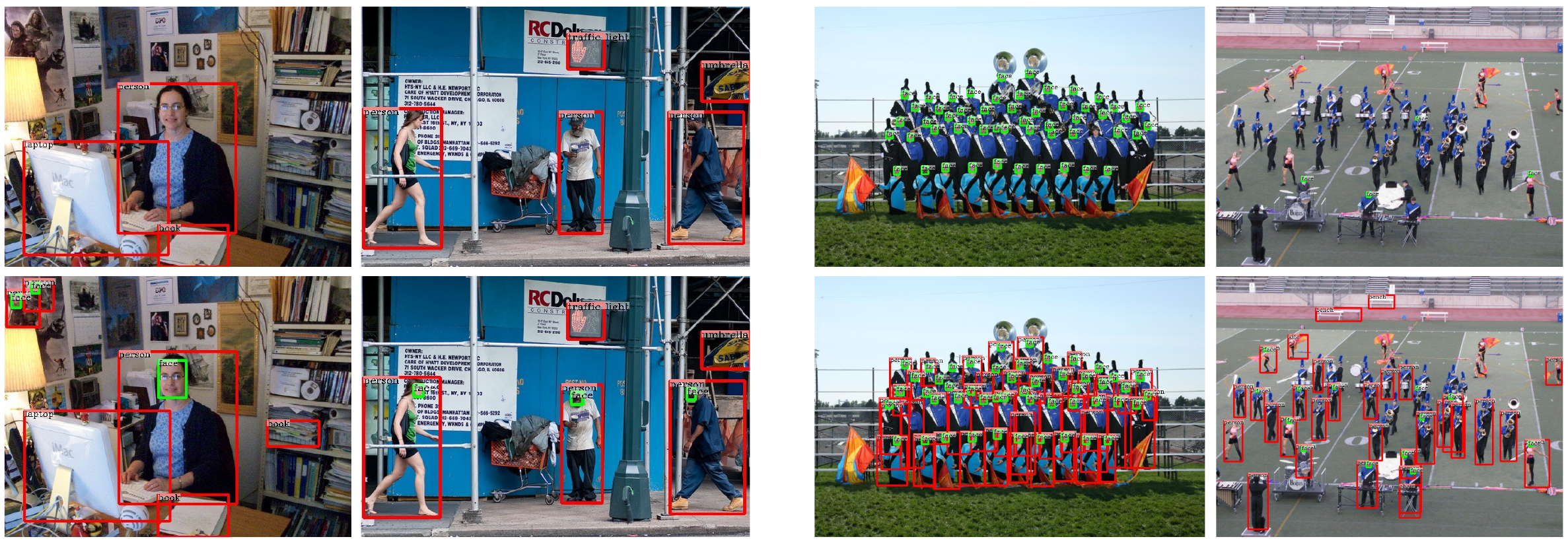}
  \caption{Detection results of the plain joint training (top) and our method (bottom) on COCO-val (COCO evaluation set, the first two columns) and FACE-val (WIDER-FACE evaluation set, the last two columns). Red bounding boxes indicate objects belonging to the 80 categories in COCO, while green ones correspond to the category ``face'' in WIDER FACE. Best seen in color.}
  \label{fig:vis1}
\end{figure}

We first show the performance of unified detectors trained by the proposed solution and the plain joint training on setup A-D. As shown in Fig.~\ref{fig:compare_baselines}, the performance of the detector obtained by plain joint training is poor, especially for categories that are not labeled in datasets. In setup C, due to the label ambiguity and domain gap, the detector trained in the plain way is heavily biased, which only gets 4.1\% AP$_{50}$ for the 20 VOC categories on COCO-val. In setup D, the performances of the detector trained in the plain way are only 3.4\% and 6.7\% in terms of AP$_{50}$ for ``face" on COCO-subval and ``person" on FACE-subval, respectively. We show that the proposed method achieves the performances of 42.6\%, 53.9\% and 58.5\% with improvements of 38.5\%, 50.5\% and 51.8\%, respectively. For setup A and B, we also provide the results based on fully labeled datasets, which can be seen as upper-bounds. These results demonstrate that our method can reduce the gap with fully labeled datasets.

Fig.~\ref{fig:vis1} illustrates some visualization results of the detectors trained on COCO-train and FACE-train. Due to the conflicts during plain joint training and the domain gap between the two sets, the detector trained in the vanilla way only can detect ``person'' in COCO-val and ``face'' in FACE-val (i.e., categories which are labeled in corresponding training data). Assisted by our approach, the detector can detect all categories on two evaluation sets. Both qualitative and quantitative results demonstrate the effectiveness of our method.

\subsection{Comparison to Other Methods}
In this section, we compare our conflict-free loss, overlap-weighted retraining method and CLC-based unlabeled ground-truth mining method with corresponding counterparts in Sec.~\ref{sec:compare_1}, Sec.~\ref{sec:compare_2} and Sec.~\ref{sec:compare_3}, respectively.

\begin{table}[t]
\centering
\caption{Compare conflict-free loss with other methods on COCO75-train ($\mathcal{D}_o$) and COCO5-train ($\mathcal{D}_n$). The results of 75 original classes and 5 new classes on COCO-val are reported. `\textbf{*}' indicates the categories that we should pay more attention to. Best results for these categories are in \textbf{bold}.}
\resizebox{1.\textwidth}{!}{
\begin{tabular}{llllllll}
\toprule
Data  & Method & \multicolumn{3}{l}{75 original classes} & \multicolumn{3}{l}{5 new classes~\textbf{*}} \\
\cmidrule(lr){3-5}  \cmidrule(lr){6-8}   
&     & AP    & $\text{AP}_{50}$   & $\text{AP}_{75}$   &  AP    & AP$_{50}$   & AP$_{75}$  \\
\midrule
$\mathcal{D}_o$   &  plain~\cite{lin2017focal}  & 36.3  & 55.4  & 38.6 & N/A    & N/A   & N/A \\
$\mathcal{D}_n$  &     plain~\cite{lin2017focal}  & N/A    & N/A    & N/A & 20.2  & 36.6  & 19.4  \\
$\mathcal{D}_o$+$\mathcal{D}_n$  &   plain~\cite{lin2017focal}  & 35.8  & 54.9  & 38.1 & 20.6  & 36.3  & 20.3  \\
  &  dataset-aware~\cite{yao2020cross}  & 36.4  & 55.3  & \textbf{38.8} & 22.7  & 39.4  & 22.5  \\
   &  conflict-free (ours)  & \textbf{36.5}  &  \textbf{55.5} & 38.7 & \textbf{23.5}  &  \textbf{40.4}  &  \textbf{23.4}  \\
{\color{highlight} fully labeled} &   {\color{highlight}plain~\cite{lin2017focal}}  & {\color{highlight}37.1}  & {\color{highlight}56.3}  & {\color{highlight}39.6} & {\color{highlight}30.6}  & {\color{highlight}49.6}  & {\color{highlight}31.5} \\
\bottomrule
\end{tabular}%
}
\label{tab:compare_loss}%
\end{table}%

\subsubsection{Compare conflict-free loss with others} \label{sec:compare_1}

We compare the proposed conflict-free loss with other methods under setup A. The dataset-aware loss in~\cite{yao2020cross} is a special case of conflict-free loss without considering the second conflict origin discussed in Sec.~\ref{sec:cf_loss}. We also provide the results of multiple detectors trained on separate datasets.

As shown in Tab.~\ref{tab:compare_loss}, the performance of the detectors trained separately on COCO75-train and COCO5-train are 36.3\% AP and 20.2\% AP respectively. The performance of the detector trained on the combined training set in a vanilla way drops to 35.8\% AP for the 75 classes, which is mainly caused by the conflicts during joint training; and improves to 20.6\% AP for the 5 classes, which reveals that more training data may lead to better feature representation. The detector trained with dataset-aware loss~\cite{yao2020cross} outperforms the plain method. Furthermore, our conflict-free loss achieves better results (23.5\% AP for the 5 classes), which is mainly due to that the conflict-free loss can avoid two possible conflicts simultaneously as discussed in Sec.~\ref{sec:cf_loss}. It is worth noting that the unified detector trained with conflict-free loss outperforms separate detectors, especially for new classes (23.5AP vs 20.2AP) owing to the abundant training samples provided by other datasets. It demonstrates that the conflict-free loss leads to a unified detector with acceptable results in one training round, outperforms other unified detectors, and is not worse than separate detectors.

\begin{table}[t]
     \centering
     \caption{Compare the unlabeled ground-truth mining method and the retraining method with other approaches under setup A ($\mathcal{D}_o$: COCO75-train, $\mathcal{D}_n$: COCO5-train). $\mathcal{\hat{G}}_{CC}$ and $\mathcal{\hat{G}}_{CLC}$ represent the mined pseudo annotations by CC-based and CLC-based methods, respectively. We report the results of the 75 original classes and the 5 new classes on COCO-val, respectively.}
\resizebox{1\textwidth}{!}{
\begin{tabular}{llllllll}
\toprule
Data & Method & \multicolumn{3}{l}{75 original classes} & \multicolumn{3}{l}{5 new classes~\textbf{*}} \\
\cmidrule(lr){3-5}  \cmidrule(lr){6-8} 
&       & AP    & $\text{AP}_{50}$ & $\text{AP}_{75}$ & AP    & AP$_{50}$ & AP$_{75}$ \\
\midrule
$\mathcal{D}_o$+$\mathcal{D}_n$ & plain~\cite{lin2017focal} & 35.8  & 54.9  & 38.1  &  20.6  &  36.3  &  20.3  \\
  & conflict-free (ours) & 36.5  & 55.5  & 38.7  &   23.5  &  40.4  &  23.4  \\
$\mathcal{D}_o$+$\mathcal{D}_n$+$\mathcal{\hat{G}}_{CC}$ & as fully labeled~\cite{sohn2020simple} & 36.6  & 55.4  & 38.9 &    24.0  &  39.9  &  24.6  \\
  & conflict-free (ours) & 36.3  & 55.2  & 38.5  &   24.2  &  40.8  &  24.6  \\
  & safe negatives~\cite{zhao2020object} & 36.6  & \textbf{55.7} & 38.7  &  24.5  &  41.1  &  25.0  \\
  & overlap-weighted (ours) & 36.6  & 55.4  & \textbf{39.0} &  24.7  &  41.7 &  \textbf{25.1}  \\
$\mathcal{D}_o$+$\mathcal{D}_n$+$\mathcal{\hat{G}}_{CLC}$    & overlap-weighted (ours) & \textbf{36.7} & 55.6  & \textbf{39.0} &  \textbf{24.9} &  \textbf{42.0} &  \textbf{25.1} \\
{\color{highlight} fully labeled} &   {\color{highlight}plain~\cite{lin2017focal}}  & {\color{highlight}37.1}  & {\color{highlight}56.3}  & {\color{highlight}39.6} & {\color{highlight}30.6}  & {\color{highlight}49.6}  & {\color{highlight}31.5} \\
\bottomrule
\end{tabular}%
}
\label{tab:coco75+coco5}%
\end{table}%

\begin{table}[t]
\centering
\caption{Compare the unlabeled ground-truth mining method and the retraining method with other approaches under setup B ($\mathcal{D}_o$: COCO79-train, $\mathcal{D}_n$: COCO1-train). We report the results of the 79 original classes and the 1 new class on COCO-val, respectively.}
\resizebox{1.\textwidth}{!}{
\begin{tabular}{llllllll}
\toprule
Data &  Method & \multicolumn{3}{l}{79 original classes} & \multicolumn{3}{c}{1 new class~\textbf{*}} \\
\cmidrule(lr){3-5}    \cmidrule(lr){6-8}  
&       & AP    & AP$_{50}$ & AP$_{75}$  & AP    & AP$_{50}$ & AP$_{75}$ \\
\midrule
$\mathcal{D}_o$+$\mathcal{D}_n$ & plain~\cite{lin2017focal} & 35.4  & 54.4  & 37.7  &  37.9  &  69.2  &  36.5  \\
 & conflict-free (ours) & \textbf{36.5}  & \textbf{55.5}  & \textbf{39.0} &  41.6  &  72.4  &  41.4  \\
$\mathcal{D}_o$+$\mathcal{D}_n$+$\mathcal{\hat{G}}_{CC}$ & as fully labeled~\cite{sohn2020simple} &   36.2    &   55.1    &   38.4    &    42.9   &    71.1   &  44.4 \\
  & conflict-free (ours) &  36.1  & 54.8  &  38.3 &  44.4 & 74.4  & 45.6  \\
 & safe negatives~\cite{zhao2020object} & 36.1  & 55.0  & 38.2  & 43.7  & 72.9  & 45.0  \\
 & overlap-weighted (ours) & 36.0  & 54.8  & 38.1 & 44.6  & 74.2  & 46.0  \\
$\mathcal{D}_o$+$\mathcal{D}_n$+$\mathcal{\hat{G}}_{CLC}$ & overlap-weighted (ours) & 36.2  & 55.0  & 38.5 & \textbf{44.7}  & \textbf{74.6}  & \textbf{46.3} \\
{\color{highlight} fully labeled} &   {\color{highlight}plain~\cite{lin2017focal}}  & {\color{highlight}36.5}  & {\color{highlight}55.5}  & {\color{highlight}38.9} & {\color{highlight}51.2}  & {\color{highlight}79.6}  & {\color{highlight}54.2} \\
\bottomrule
\end{tabular}%
}
\label{tab:coco79+coco1}
\end{table}%

\subsubsection{Compare the overlap-weighted retraining method with others} 
\label{sec:compare_2}

We describe other retraining strategies as follows. (i) As described in Sec.~\ref{sec:retraining}, we can view the combined dataset supplemented with pseudo annotations \textbf{as fully labeled}~\cite{sohn2020simple}, and train detectors with common classification loss. (ii) We can retrain the detector with the \textbf{conflict-free} loss again. Compared to the first phase, only positive samples are supplemented by the pseudo annotations. (iii) We can introduce some \textbf{safe negatives}~\cite{zhao2020object}. In addition to the threshold $\eta$ used to mine high-quality pseudo annotations, a lower threshold $\eta'$ is introduced to get high-recall pseudo annotations. Under the combination of ground-truth and high-quality pseudo annotations, some negative samples are actually unsafe negative samples which can match boxes from high-recall pseudo annotations. Anchors that do not match any ground-truth or high-recall pseudo annotations are considered to be safe negative samples. Safe negative samples can contribute to the loss of classes from different datasets while unsafe negative samples are still discarded.

For fair, we compare our overlap-weighted retraining method with these strategies with the same pseudo annotations (training detectors with conflict-free loss and mining pseudo annotations with classification confidence, which is described in the next paragraph). We provide the results of setup A, B, C and D in Tab.~\ref{tab:coco75+coco5},~\ref{tab:coco79+coco1},~\ref{tab:coco60+voc20} and~\ref{tab:coco80+face1} respectively ($\mathcal{D}_o$ + $\mathcal{D}_n$ + $\mathcal{\hat{G}}_{CC}$). In setup A, retraining with ``as fully labeled" leads to improvement compared to the results of one training round (0.5\%AP gains for the 5 new classes), which demonstrates the effectiveness of retraining. The detector retrained with conflict-free loss achieves better performance for the 5 classes. Relying on ``safe negatives" during retraining, the detector reaches 24.5\% AP for the 5 classes. Finally, our retraining method achieves the best results (24.7\% AP for 5 classes). Similar results are shown in setup B, C and D. These results imply that there are still many unlabeled ground-truth in the training data, thus the idea of avoiding conflicts cannot be ignored in retraining process. Whereas, if we use the same method as in the first training round, only positive samples would be added, causing insufficient negative information in retraining. Thus, it is important to employ suitable negative samples to achieve a balance between positive and negative samples. Our empirical experiments show that using the overlap-weighted negative samples is a better choice for the purpose.

\begin{table}[t]
\centering
\caption{Compare the unlabeled ground-truth mining method and the retraining method with other approaches under setup C ($\mathcal{D}_o$: COCO60-train, $\mathcal{D}_n$: VOC-train). We report the results of all 80 classes on VOC-subtest and COCO-subval, and the 20 VOC classes on COCO-val.
}
\resizebox{1.\textwidth}{!}{
\begin{tabular}{llllllll}
\toprule
Data & Method & \multicolumn{2}{l}{VOC-subtest}  & \multicolumn{2}{l}{COCO-subval} & \multicolumn{2}{l}{COCO-val} \\
\cmidrule(lr){3-4}      \cmidrule(lr){5-6}      \cmidrule(lr){7-8}      
  &   & \multicolumn{2}{l}{all 80 classes} & \multicolumn{2}{l}{all 80 classes} & \multicolumn{2}{l}{20 new classes} \\
\cmidrule(lr){3-4}      \cmidrule(lr){5-6}      \cmidrule(lr){7-8}      
&       & AP$_{50}$ & AP$_{75}$ & AP$_{50}$ & AP$_{75}$ & AP$_{50}$ & AP$_{75}$ \\
\midrule
$\mathcal{D}_o$+$\mathcal{D}_n$ & plain~\cite{lin2017focal} & 43.0  & 28.0 & 42.6  & 27.2  & 4.1   & 2.7  \\
 & conflict-free (ours) & 46.8  & 30.1 &  50.3  & 30.8  & 32.3  & 18.6  \\
$\mathcal{D}_o$+$\mathcal{D}_n$+$\mathcal{\hat{G}}_{CC}$ & as fully labeled~\cite{sohn2020simple} &  48.0     &   31.5  &  48.7    &    29.9 &   27.5    & 18.6 \\
    & conflict-free (ours) &  \textbf{49.4}  & 31.7 &  49.9 & 30.8 & 34.1 & 20.4 \\
    & safe negatives~\cite{zhao2020object} & 49.0  &  \textbf{33.3}  & 49.9  & 31.5 & 34.3  & 21.7  \\
 & overlap-weighted (ours) & 47.5  & 31.4 & 51.6  & 31.6  & \textbf{42.7}  & 24.9  \\
$\mathcal{D}_o$+$\mathcal{D}_n$+$\mathcal{\hat{G}}_{CLC}$ & overlap-weighted (ours) & 49.0  & 33.0 & \textbf{52.2}  & \textbf{31.7}   & 42.6  & \textbf{25.4}  \\
\bottomrule
\end{tabular}%
}
\label{tab:coco60+voc20}%
\end{table}%

\begin{table}[t]
\centering
\caption{Compare the unlabeled ground-truth mining method and the retraining method with other approaches on setup D ($\mathcal{D}_o$: COCO-train, $\mathcal{D}_n$: FACE-train). We report the results of categories ``person'' and ``face''. `\textbf{*}' indicates the categories that we should pay more attention to.}
\resizebox{1.\textwidth}{!}{
\begin{tabular}{llllllllll}
\toprule
Data & Method & \multicolumn{4}{l}{COCO-subval}  & \multicolumn{4}{l}{FACE-subval} \\
\cmidrule(lr){3-6}     \cmidrule(lr){7-10} 
   &   & \multicolumn{2}{l}{person} & \multicolumn{2}{l}{face~\textbf{*}} &  \multicolumn{2}{l}{person~\textbf{*}} & \multicolumn{2}{l}{face} \\
\cmidrule(lr){3-4} \cmidrule(lr){5-6} \cmidrule(lr){7-8} \cmidrule(lr){9-10} 
   &   & AP$_{50}$ & AP$_{75}$ & AP$_{50}$ & AP$_{75}$ & AP$_{50}$ & AP$_{75}$ & AP$_{50}$ & AP$_{75}$ \\
\midrule
$\mathcal{D}_o$+$\mathcal{D}_n$ & plain~\cite{lin2017focal} & 71.1  & 40.8  &  3.4   &  0.9   &  6.7   &  3.7   & 54.3  & 28.0  \\
& conflict-free (ours) & 71.9  & \textbf{43.2}  &  49.9  &  20.9  &  55.6  &  33.9  & \textbf{53.8}  & 28.8  \\
$\mathcal{D}_o$+$\mathcal{D}_n$+$\mathcal{\hat{G}}_{CC}$ & as fully labeled~\cite{sohn2020simple} &  \textbf{72.7}   & 42.6      &  49.0      &   22.6    &  51.5     &    31.8    &    \textbf{53.8}   & 28.9 \\
& conflict-free (ours) & 71.2  &  42.1  &  53.9  &   22.8 &  56.4   &    \textbf{34.7}   &  53.7  & 28.8 \\
& safe negatives~\cite{zhao2020object} & 72.4 & 42.3  &  50.7  &  22.4 &  52.9  &  31.7  & 54.2  & \textbf{29.0}  \\
& overlap-weighted (ours) & 71.4  & 42.8  &  \textbf{54.0}  &  22.5  &  58.3  &  \textbf{34.7}  & 53.1  & 28.7  \\
$\mathcal{D}_o$+$\mathcal{D}_n$+$\mathcal{\hat{G}}_{CLC}$ & overlap-weighted (ours) & 72.3  & 43.1  &  53.9  &  \textbf{23.7} &  \textbf{58.5} &  \textbf{34.7} & 53.2  & \textbf{29.0}  \\
\bottomrule
\end{tabular}%
}
\label{tab:coco80+face1}%
\end{table}%

\begin{figure}[t]
  \centering
  \includegraphics[height=0.12\textheight]{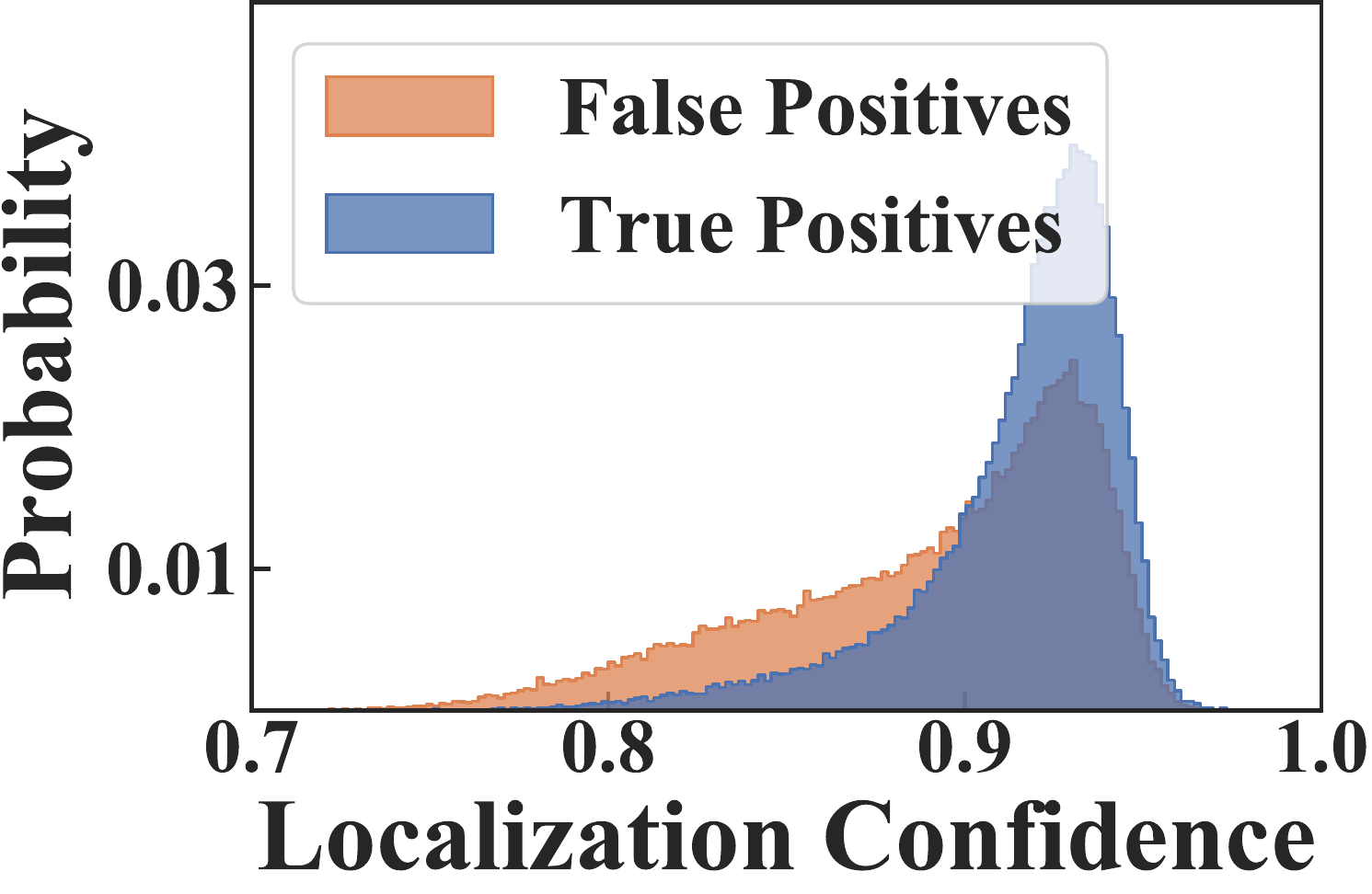}
  \includegraphics[height=0.12\textheight]{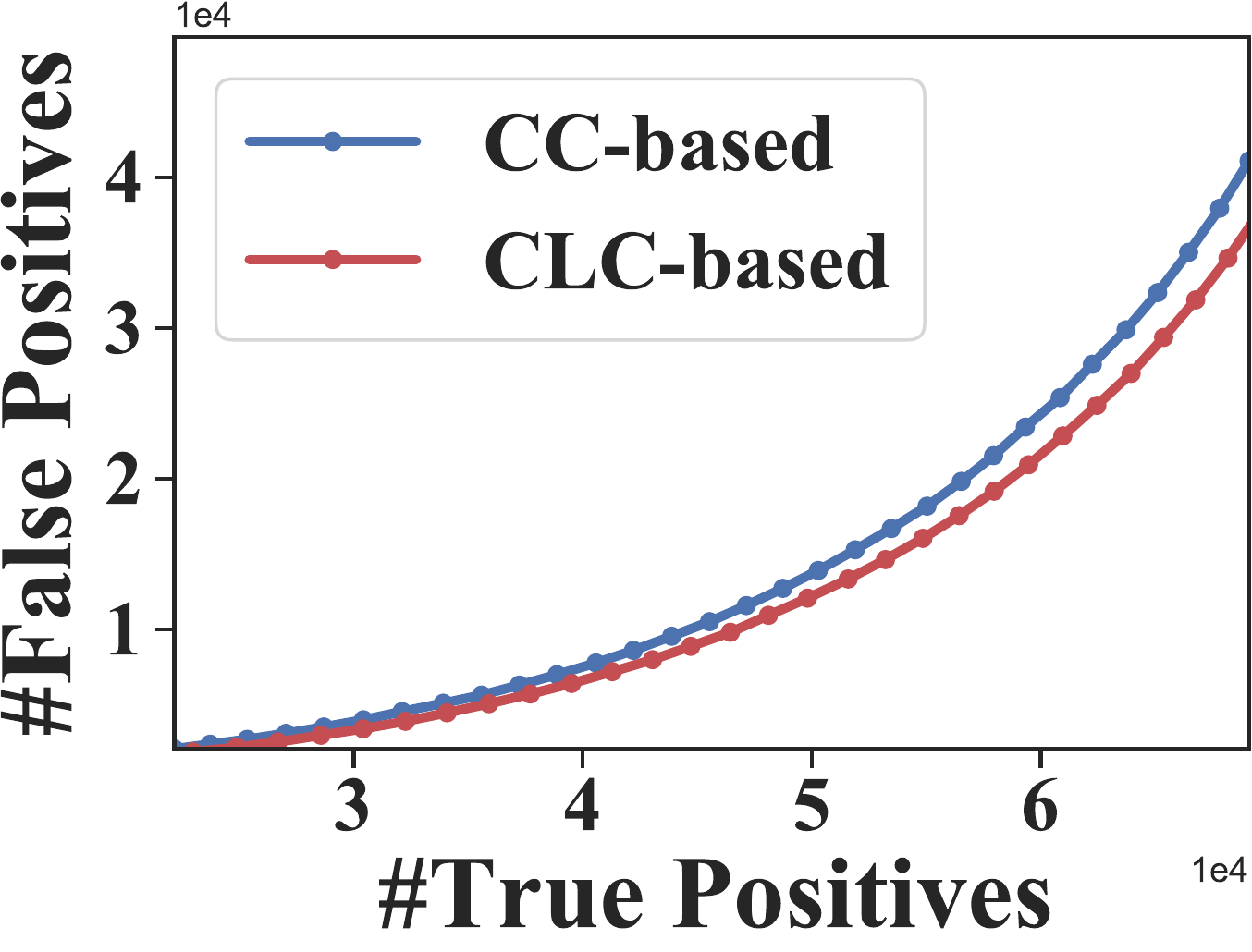}
  \caption{Distribution of localization confidence across true and false positive predictions (left). The curve of true positive predictions versus fasle negative predictions (right). Best seen in color.}
  \label{fig:tpfp}
\end{figure}

\subsubsection{Compare CLC-based unlabeled ground-truth mining method with others}
\label{sec:compare_3}

We compare the CLC-based mining method with the approach that only relies on classification confidence (CC-based) similar to some work for classification tasks~\cite{berthelot2019mixmatch,huang2021behavior,lai2019improving}. The generating process also relies on a score threshold strategy, which only exploits the classification score to represent the detector's confidence in the predicted results. The predictions with high classification confidence will be selected as pseudo annotations to complete ground-truth for the unlabeled classes.

We investigate the performance of the CC-based and the CLC-based mining methods. We expect more accurate bounding boxes can be mined by our CLC-based mining method. To check the localization quality of bounding boxes, we count the true positive predictions and the false positive predictions with high overlap metric ($\text{IoU} \geq 0.75$) in the predicted results whose classification confidence is larger than 0.5. The distribution of localization confidence (Eq.~\eqref{eq:loc_conf}) across true positive predictions and false positive predictions are shown in Fig.~\ref{fig:tpfp} (left). We find the overwhelming majority of bounding boxes with low localization confidence are false negative predictions, which shows the localization confidence can reflect the quality of bounding boxes to some extent. Besides, as shown in Fig.~\ref{fig:tpfp} (right), the number of false positive predictions can be suppressed with the collaboration of localization confidence. The results in Table~\ref{tab:coco75+coco5},~\ref{tab:coco79+coco1},~\ref{tab:coco60+voc20} and~\ref{tab:coco80+face1} ($\mathcal{D}_o$ + $\mathcal{D}_n$ + $\mathcal{\hat{G}}_{CLC}$) also demonstrate that the pseudo annotations mined by the CLC-based method bring an improvement compared to the CC-based method.

\subsection{Extend to multiple training sets}
The above analysis and evaluations are conducted on two datasets, while, our method is also fully capable of handling multiple datasets. To verify it, we perform setup E, which involves three training sets labeled with different classes: COCO60-train ($\mathcal{D}_o$), VOC-train ($\mathcal{D}_{n_1}$), and FACE-train ($\mathcal{D}_{n_2}$). The results are listed in Table~\ref{tab:setup_e}, demonstrating the effectiveness of our solution again.
\begin{table}[tp]
\centering
\caption{Results on setup E. AP$_{50}$ (\%) / AP$_{75}$ (\%) are reported for the 81 classes on VOC-subtest, the 81 classes on COCO-subval, and the ``person" category on FACE-subval, respectively.}
\resizebox{1.\textwidth}{!}{
\begin{tabular}{llllllll}
\toprule
Data & Method & \multicolumn{2}{l}{VOC-subtest}  & \multicolumn{2}{l}{COCO-subval}  & \multicolumn{2}{l}{FACE-subval} \\
\cmidrule(lr){3-4} \cmidrule(lr){5-6} \cmidrule(lr){7-8}
  &   & \multicolumn{2}{l}{all 81 classes}  & \multicolumn{2}{l}{ all 81 classes}  & \multicolumn{2}{l}{person} \\
\cmidrule(lr){3-4} \cmidrule(lr){5-6} \cmidrule(lr){7-8}
& &  AP$_{50}$ &  AP$_{75}$  & AP$_{50}$  & AP$_{75}$  & AP$_{50}$  &  AP$_{75}$\\
\midrule
$\mathcal{D}_o$+$\mathcal{D}_{n_1}$+$\mathcal{D}_{n_2}$ & plain~\cite{lin2017focal} & 40.0 & 27.0 & 42.1 & 25.7 & 0.2 & 0.0 \\
  & conflict-free (ours) & 46.5 & 30.1 & 49.1 & 29.7 & 41.9 & 13.5 \\
$\mathcal{D}_o$+$\mathcal{D}_{n_1}$+$\mathcal{D}_{n_2}$+$\mathcal{\hat{G}}_{CLC}$ & as fully labeled~\cite{sohn2020simple} & 47.4 & \textbf{32.0} & 48.2 & 30.0 & 39.3 & 15.7 \\
 & overlap-weighted (ours) & \textbf{49.6} & 31.4 & \textbf{51.2} & \textbf{30.4} & \textbf{48.5} & \textbf{18.5} \\
\bottomrule
\end{tabular}%
}
\label{tab:setup_e}%}%
\end{table}%

\subsection{BCE-based v.s. CE-based}
Similar to BCE loss, cross-entropy (CE) loss is also widely used in object detection. With CE loss, label $\mathbf{p}^*$ and prediction $\mathbf{p}$ should have $|\mathcal{C}_o \cup \mathcal{C}_n|+1$ categories (plus the ``background"), and the prediction $\mathbf{p}$ is activated by softmax instead of sigmoid. BCE-based and CE-based loss are both commonly used. There is usually no obvious difference in performance between these two losses in standard object detection~\cite{redmon2018yolov3}. However, in CE loss, all categories would affect each other (due to ``softmax"). Thus, we argue that BCE-based loss is a better choice for the task handled in this work, as it helps to avoid conflicts conveniently.

\begin{table}[t]
\centering
\caption{{\color{highlight}Performance with pseudo annotations mined by our unified detector in Sec.~\ref{sec:cf_loss} ($\mathcal{\hat{G}}_{CC}$, $\mathcal{\hat{G}}_{CLC}$) and the Separate Detectors ($\mathcal{\hat{G}}_{CC,SDs}$, $\mathcal{\hat{G}}_{CLC,SDs}$) under setup C.}}
\resizebox{1.\textwidth}{!}{
\begin{tabular}{lllllll}
\toprule
\multicolumn{1}{l}{\color{highlight}{Method}} & \color{highlight}{$\mathcal{\hat{G}}$}  & \multicolumn{2}{l}{\color{highlight}{VOC-subtest}} & \color{highlight}{} & \multicolumn{2}{l}{\color{highlight}{COCO-subval}} \\
\cmidrule{3-4}\cmidrule{6-7}\color{highlight}{} &  & \color{highlight}{AP$_{50}$} & \color{highlight}{AP$_{75}$} & \color{highlight}{} & \color{highlight}{AP$_{50}$} & \color{highlight}{AP$_{75}$} \\
\midrule
\multicolumn{1}{l}{\color{highlight}{as fully labeled~\cite{sohn2020simple}}} & \color{highlight}{$\mathcal{\hat{G}}_{CC,SDs}$} & \color{highlight}{44.3 } & \color{highlight}{28.3 } & \color{highlight}{} & \color{highlight}{48.0 } & \color{highlight}{27.2 } \\
\color{highlight}{} & \color{highlight}{$\mathcal{\hat{G}}_{CC}$} & \color{highlight}{48.0 (+3.7) } & \color{highlight}{31.5 (+3.2)} & \color{highlight}{} & \color{highlight}{48.7 (+0.7)} & \color{highlight}{29.9 (+2.7)} \\
\multicolumn{1}{l}{\color{highlight}{confict-free}} & \color{highlight}{$\mathcal{\hat{G}}_{CC,SDs}$} & \color{highlight}{46.2 } & \color{highlight}{26.8 } & \color{highlight}{} & \color{highlight}{50.1 } & \color{highlight}{29.9 } \\
\color{highlight}{} & \color{highlight}{$\mathcal{\hat{G}}_{CC}$} & \color{highlight}{49.4 (+3.2)} & \color{highlight}{31.7 (+4.9)} & \color{highlight}{} & \color{highlight}{49.9 (-0.2)} & \color{highlight}{30.8 (+0.9)} \\
\multicolumn{1}{l}{\color{highlight}{safe negatives~\cite{zhao2020object}}} & \color{highlight}{$\mathcal{\hat{G}}_{CC,SDs}$} & \color{highlight}{45.3 } & \color{highlight}{28.2 } & \color{highlight}{} & \color{highlight}{49.7 } & \color{highlight}{28.2 } \\
\color{highlight}{} & \color{highlight}{$\mathcal{\hat{G}}_{CC}$} & \color{highlight}{49.0 (+3.7)} & \color{highlight}{33.3 (+5.1)} & \color{highlight}{} & \color{highlight}{49.9 (+0.2)} & \color{highlight}{31.5 (+3.3)} \\
\multicolumn{1}{l}{\color{highlight}{overlap-weighted}} & \color{highlight}{$\mathcal{\hat{G}}_{CC,SDs}$} & \color{highlight}{46.3 } & \color{highlight}{28.7 } & \color{highlight}{} & \color{highlight}{50.0 } & \color{highlight}{29.4 } \\
\color{highlight}{} & \color{highlight}{$\mathcal{\hat{G}}_{CC}$} & \color{highlight}{47.5 (+1.2)} & \color{highlight}{31.4 (+2.7)} & \color{highlight}{} & \color{highlight}{51.6 (+1.6)} & \color{highlight}{31.6 (+2.2)} \\
\multicolumn{1}{l}{\color{highlight}{overlap-weighted}} & \color{highlight}{$\mathcal{\hat{G}}_{CLC,SDs}$} & \color{highlight}{49.2 } & \color{highlight}{30.8 } & \color{highlight}{} & \color{highlight}{50.9 } & \color{highlight}{29.9 } \\
\color{highlight}{} & \color{highlight}{$\mathcal{\hat{G}}_{CLC}$} & \color{highlight}{49.0 (-0.2)} & \color{highlight}{33.0 (+2.2)} & \color{highlight}{} & \color{highlight}{52.2 (+1.3)} & \color{highlight}{31.7 (+1.8)} \\
\bottomrule
\end{tabular}%
}
\label{tab:sds}%
\end{table}%

{\color{highlight}
\subsection{Separate detectors for unlabeled ground-truth mining}
\label{sec:SDs}
We have shown that the conflict-free loss can lead to an acceptable detector in one training round, and that the retraining phase can significantly improve performance. Conflict-free loss is recommended if you prefer a simple one-round training approach. If one adopts the two-rounds training, what about using multiple detectors trained on separate datasets for unlabeled ground-truth mining? 

We compare the performance of using pseudo annotations mined by the unified detector and the separate detectors under setup C. The results in Tab.~\ref{tab:sds} demonstrate the pseudo annotations mined by the unified model are superior to those mined by multiple separate detectors. Considering that training multiple separate detectors is more cumbersome, we also recommend the unified detector as an initial detector in two-rounds training method.
}

\subsection{More visualization results.}
\begin{figure}[tp]
\centering
\caption{Detection results on COCO-val and VOC-test. Red bounding boxes indicate objects of the 60 categories in COCO60-train, while green ones correspond to the 20 VOC categories.}
\includegraphics[height=0.6\textheight]{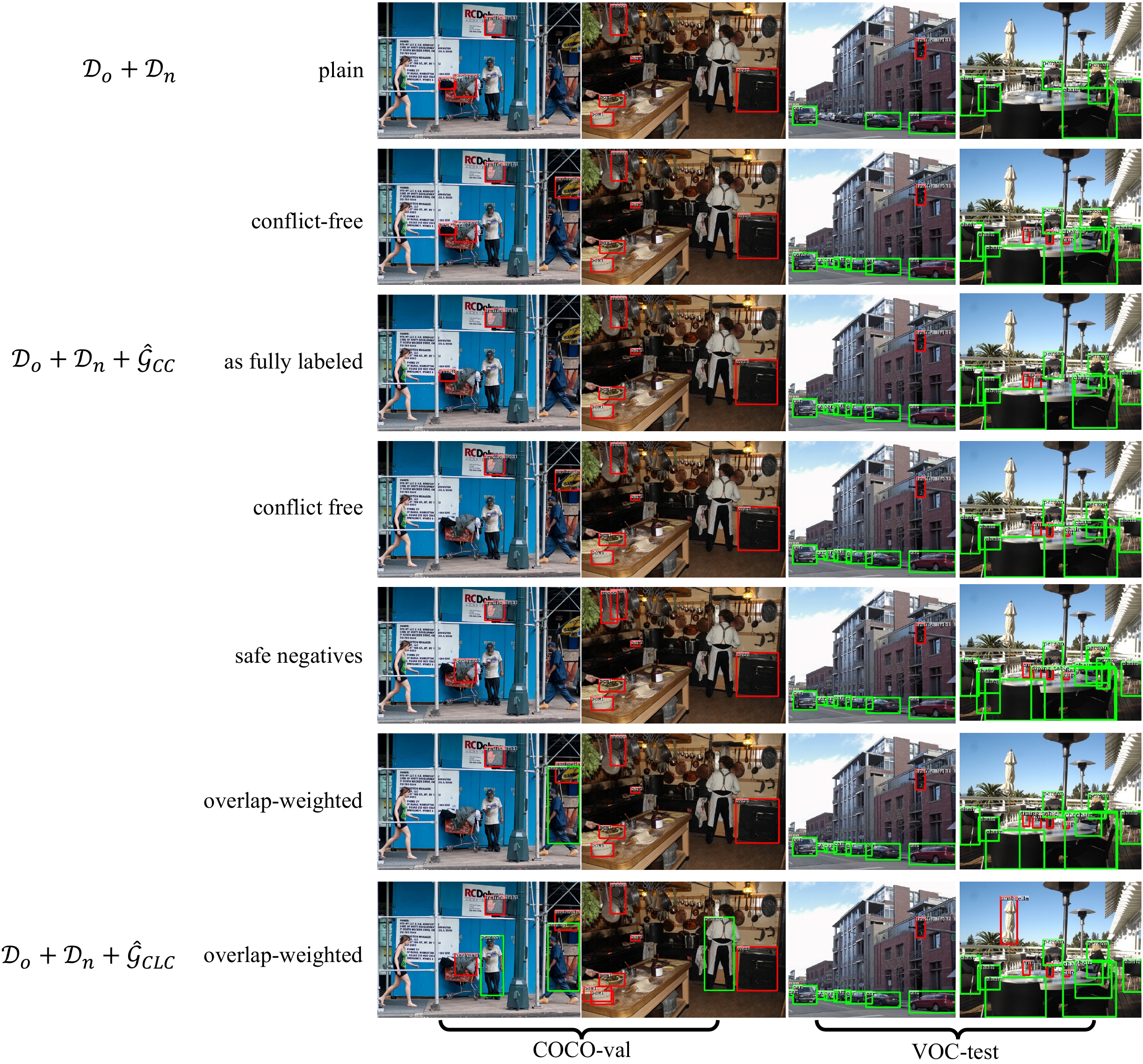}
\label{fig:vis2}%
\end{figure}

Fig.~\ref{fig:vis2} illustrates more visualization results of the detectors on setup C, showing that our method achieves powerful and robust detectors with limited data.

\section{Conclusion}
In this paper, we aim at training a category-extended object detector with limited data. To achieve this goal, a powerful pipeline is proposed. The conflict-free loss is designed to eliminate possible conflicts during joint learning. To further improve performance, a retraining phase is employed. We adopt a Monte Carlo Dropout-based method to estimate localization confidence to unearth more accurate pseudo annotations. We explore several strategies for retraining with the pseudo annotations, and empirically exhibit that employing overlap-weighted negative samples during retraining leads to better performance. The experimental results on multiple settings demonstrate that the proposed method achieves a feasible category-extended object detector and outperforms other approaches.

{\color{highlight} Despite the promising results, there is still a significant gap between models trained on incrementally labeled datasets and models trained on fully labeled datasets. Exploring unsupervised domain adaptation algorithms would benefit unlabeled ground-truth mining by mitigating the effects of domain gap; designing class-distribution aware retraining methods would be beneficial in alleviating the class imbalance problem, which may be exacerbated by pseudo annotations. For other tasks, the incrementally labeled datasets can also be encountered in multi-label classification, semantic / instance segmentation, and other modalities, which are worth exploring further.}

\section*{Acknowledgements}
This work is supported in part by the National Natural Science Foundation of China (62171248, 61972219), the R\&D Program of Shenzhen (JCYJ2018050815\\2204044, JCYJ20190813174403598, SGDX20190918101201696), the PCNL KEY project (PCL2021A07), and the Overseas Research Cooperation Fund of Tsinghua Shenzhen International Graduate School (HW2021013).



% \newpage
\begin{small}
\bibliographystyle{elsarticle-num} 
\bibliography{egbib.bib}

\begin{thebibliography}{10}
\expandafter\ifx\csname url\endcsname\relax
  \def\url#1{\texttt{#1}}\fi
\expandafter\ifx\csname urlprefix\endcsname\relax\def\urlprefix{URL }\fi
\expandafter\ifx\csname href\endcsname\relax
  \def\href#1#2{#2} \def\path#1{#1}\fi

\bibitem{WEI2020107195}
X.~Wei, H.~Zhang, S.~Liu, Y.~Lu, Pedestrian detection in underground mines via
  parallel feature transfer network, Pattern Recognition 103 (2020) 107195.

\bibitem{wang2020focalmix}
D.~Wang, Y.~Zhang, K.~Zhang, L.~Wang, Focalmix: Semi-supervised learning for 3d
  medical image detection, in: Proceedings of the IEEE/CVF Conference on
  Computer Vision and Pattern Recognition, 2020, pp. 3951--3960.

\bibitem{AKCAY2022108245}
Towards automatic threat detection: A survey of advances of deep learning
  within x-ray security imaging, Pattern Recognition 122 (2022) 108245.

\bibitem{ren2015faster}
S.~Ren, K.~He, R.~Girshick, J.~Sun, Faster r-cnn: Towards real-time object
  detection with region proposal networks, in: Advances in neural information
  processing systems, 2015, pp. 91--99.

\bibitem{lin2017focal}
T.-Y. Lin, P.~Goyal, R.~Girshick, K.~He, P.~Doll{\'a}r, Focal loss for dense
  object detection, in: Proceedings of the IEEE international conference on
  computer vision, 2017, pp. 2980--2988.

\bibitem{QIN2020107404}
X.~Qin, Z.~Zhang, C.~Huang, M.~Dehghan, O.~R. Zaiane, M.~Jagersand, U2-net:
  Going deeper with nested u-structure for salient object detection, Pattern
  Recognition 106 (2020) 107404.

\bibitem{lin2014microsoft}
T.-Y. Lin, M.~Maire, S.~Belongie, J.~Hays, P.~Perona, D.~Ramanan,
  P.~Doll{\'a}r, C.~L. Zitnick, Microsoft coco: Common objects in context, in:
  European conference on computer vision, Springer, 2014, pp. 740--755.

\bibitem{yang2016wider}
S.~Yang, P.~Luo, C.-C. Loy, X.~Tang, Wider face: A face detection benchmark,
  in: Proceedings of the IEEE conference on computer vision and pattern
  recognition, 2016, pp. 5525--5533.

\bibitem{redmon2016you}
J.~Redmon, S.~Divvala, R.~Girshick, A.~Farhadi, You only look once: Unified,
  real-time object detection, in: Proceedings of the IEEE conference on
  computer vision and pattern recognition, 2016, pp. 779--788.

\bibitem{yang2020distilling}
Y.~Yang, J.~Qiu, M.~Song, D.~Tao, X.~Wang, Distilling knowledge from graph
  convolutional networks, in: Proceedings of the IEEE/CVF Conference on
  Computer Vision and Pattern Recognition, 2020, pp. 7074--7083.

\bibitem{yang2020factorizable}
Y.~Yang, Z.~Feng, M.~Song, X.~Wang, Factorizable graph convolutional networks,
  Advances in Neural Information Processing Systems 33 (2020) 20286--20296.

\bibitem{jing2020dynamic}
Y.~Jing, X.~Liu, Y.~Ding, X.~Wang, E.~Ding, M.~Song, S.~Wen, Dynamic instance
  normalization for arbitrary style transfer, in: Proceedings of the AAAI
  Conference on Artificial Intelligence, Vol.~34, 2020, pp. 4369--4376.

\bibitem{jing2021amalgamating}
Y.~Jing, Y.~Yang, X.~Wang, M.~Song, D.~Tao, Amalgamating knowledge from
  heterogeneous graph neural networks, in: Proceedings of the IEEE/CVF
  Conference on Computer Vision and Pattern Recognition, 2021, pp.
  15709--15718.

\bibitem{girshick2014rich}
R.~Girshick, J.~Donahue, T.~Darrell, J.~Malik, Rich feature hierarchies for
  accurate object detection and semantic segmentation, in: Proceedings of the
  IEEE conference on computer vision and pattern recognition, 2014, pp.
  580--587.

\bibitem{girshick2015fast}
R.~Girshick, Fast r-cnn, in: Proceedings of the IEEE international conference
  on computer vision, 2015, pp. 1440--1448.

\bibitem{tian2019fcos}
Z.~Tian, C.~Shen, H.~Chen, T.~He, Fcos: Fully convolutional one-stage object
  detection, in: Proceedings of the IEEE international conference on computer
  vision, 2019, pp. 9627--9636.

\bibitem{wang2018geometry}
F.~Wang, L.~Zhao, X.~Li, X.~Wang, D.~Tao, Geometry-aware scene text detection
  with instance transformation network, in: Proceedings of the IEEE Conference
  on Computer Vision and Pattern Recognition, 2018, pp. 1381--1389.

\bibitem{yang2021training}
Z.~Yang, M.~Shi, C.~Xu, V.~Ferrari, Y.~Avrithis, Training object detectors from
  few weakly-labeled and many unlabeled images, Pattern Recognition 120 (2021)
  108164.

\bibitem{gao2022discrepant}
W.~Gao, F.~Wan, J.~Yue, S.~Xu, Q.~Ye, Discrepant multiple instance learning for
  weakly supervised object detection, Pattern Recognition 122 (2022) 108233.

\bibitem{jeong2019consistency}
J.~Jeong, S.~Lee, J.~Kim, N.~Kwak, Consistency-based semi-supervised learning
  for object detection, in: Advances in neural information processing systems,
  2019, pp. 10759--10768.

\bibitem{sohn2020simple}
K.~Sohn, Z.~Zhang, C.-L. Li, H.~Zhang, C.-Y. Lee, T.~Pfister, A simple
  semi-supervised learning framework for object detection, arXiv preprint
  arXiv:2005.04757 (2020).

\bibitem{hall2020probabilistic}
D.~Hall, F.~Dayoub, J.~Skinner, H.~Zhang, D.~Miller, P.~Corke, G.~Carneiro,
  A.~Angelova, N.~S{\"u}nderhauf, Probabilistic object detection: Definition
  and evaluation, in: The IEEE Winter Conference on Applications of Computer
  Vision, 2020, pp. 1031--1040.

\bibitem{miller2018dropout}
D.~Miller, L.~Nicholson, F.~Dayoub, N.~S{\"u}nderhauf, Dropout sampling for
  robust object detection in open-set conditions, in: 2018 IEEE International
  Conference on Robotics and Automation (ICRA), IEEE, 2018, pp. 1--7.

\bibitem{rebuffi2017icarl}
S.-A. Rebuffi, A.~Kolesnikov, G.~Sperl, C.~H. Lampert, icarl: Incremental
  classifier and representation learning, in: Proceedings of the IEEE
  conference on Computer Vision and Pattern Recognition, 2017, pp. 2001--2010.

\bibitem{wu2019large}
Y.~Wu, Y.~Chen, L.~Wang, Y.~Ye, Z.~Liu, Y.~Guo, Y.~Fu, Large scale incremental
  learning, in: Proceedings of the IEEE/CVF Conference on Computer Vision and
  Pattern Recognition, 2019, pp. 374--382.

\bibitem{zhao2020maintaining}
B.~Zhao, X.~Xiao, G.~Gan, B.~Zhang, S.-T. Xia, Maintaining discrimination and
  fairness in class incremental learning, in: Proceedings of the IEEE/CVF
  Conference on Computer Vision and Pattern Recognition, 2020, pp.
  13208--13217.

\bibitem{zhao2022energy}
B.~Zhao, C.~Chen, X.~Xiao, Q.~Ju, S.~Xia, Energy alignment for bias
  rectification in class incremental learning, in: ICASSP 2022-2022 IEEE
  International Conference on Acoustics, Speech and Signal Processing (ICASSP),
  IEEE, 2022, pp. 3513--3517.

\bibitem{michieli2021knowledge}
U.~Michieli, P.~Zanuttigh, Knowledge distillation for incremental learning in
  semantic segmentation, Computer Vision and Image Understanding 205 (2021)
  103167.

\bibitem{shmelkov2017incremental}
K.~Shmelkov, C.~Schmid, K.~Alahari, Incremental learning of object detectors
  without catastrophic forgetting, in: Proceedings of the IEEE International
  Conference on Computer Vision, 2017, pp. 3400--3409.

\bibitem{hao2019end}
Y.~Hao, Y.~Fu, Y.-G. Jiang, Q.~Tian, An end-to-end architecture for
  class-incremental object detection with knowledge distillation, in: 2019 IEEE
  International Conference on Multimedia and Expo (ICME), IEEE, 2019, pp. 1--6.

\bibitem{perez2020incremental}
J.-M. Perez-Rua, X.~Zhu, T.~M. Hospedales, T.~Xiang, Incremental few-shot
  object detection, in: Proceedings of the IEEE/CVF Conference on Computer
  Vision and Pattern Recognition, 2020, pp. 13846--13855.

\bibitem{kj2021incremental}
J.~Kj, J.~Rajasegaran, S.~Khan, F.~S. Khan, V.~N. Balasubramanian, Incremental
  object detection via meta-learning, IEEE Transactions on Pattern Analysis and
  Machine Intelligence (2021).

\bibitem{mccloskey1989catastrophic}
M.~McCloskey, N.~J. Cohen, Catastrophic interference in connectionist networks:
  The sequential learning problem, in: Psychology of learning and motivation,
  Vol.~24, Elsevier, 1989, pp. 109--165.

\bibitem{lao2021focl}
Q.~Lao, M.~Mortazavi, M.~Tahaei, F.~Dutil, T.~Fevens, M.~Havaei, Focl:
  Feature-oriented continual learning for generative models, Pattern
  Recognition 120 (2021) 108127.

\bibitem{zhao2020object}
X.~Zhao, S.~Schulter, G.~Sharma, Y.-H. Tsai, M.~Chandraker, Y.~Wu, Object
  detection with a unified label space from multiple datasets, in: European
  Conference on Computer Vision, Springer, 2020, pp. 178--193.

\bibitem{rame2018omnia}
A.~Rame, E.~Garreau, H.~Ben-Younes, C.~Ollion, Omnia faster r-cnn: Detection in
  the wild through dataset merging and soft distillation, arXiv preprint
  arXiv:1812.02611 (2018).

\bibitem{yao2020cross}
Y.~Yao, Y.~Wang, Y.~Guo, J.~Lin, H.~Qin, J.~Yan, Cross-dataset training for
  class increasing object detection, arXiv preprint arXiv:2001.04621 (2020).

\bibitem{he2016deep}
K.~He, X.~Zhang, S.~Ren, J.~Sun, Deep residual learning for image recognition,
  in: Proceedings of the IEEE conference on computer vision and pattern
  recognition, 2016, pp. 770--778.

\bibitem{jiang2018acquisition}
B.~Jiang, R.~Luo, J.~Mao, T.~Xiao, Y.~Jiang, Acquisition of localization
  confidence for accurate object detection, in: Proceedings of the European
  Conference on Computer Vision (ECCV), 2018, pp. 784--799.

\bibitem{he2019bounding}
Y.~He, C.~Zhu, J.~Wang, M.~Savvides, X.~Zhang, Bounding box regression with
  uncertainty for accurate object detection, in: Proceedings of the IEEE
  Conference on Computer Vision and Pattern Recognition, 2019, pp. 2888--2897.

\bibitem{gal2016dropout}
Y.~Gal, Z.~Ghahramani, Dropout as a bayesian approximation: Representing model
  uncertainty in deep learning, in: international conference on machine
  learning, 2016, pp. 1050--1059.

\bibitem{kendall2017uncertainties}
A.~Kendall, Y.~Gal, What uncertainties do we need in bayesian deep learning for
  computer vision?, in: Advances in neural information processing systems,
  2017, pp. 5574--5584.

\bibitem{miller2019evaluating}
D.~Miller, F.~Dayoub, M.~Milford, N.~S{\"u}nderhauf, Evaluating merging
  strategies for sampling-based uncertainty techniques in object detection, in:
  2019 International Conference on Robotics and Automation (ICRA), IEEE, 2019,
  pp. 2348--2354.

\bibitem{wu2018soft}
Z.~Wu, N.~Bodla, B.~Singh, M.~Najibi, R.~Chellappa, L.~S. Davis, Soft sampling
  for robust object detection, in: BMVC, 2019.

\bibitem{everingham2010pascal}
M.~Everingham, L.~Van~Gool, C.~K. Williams, J.~Winn, A.~Zisserman, The pascal
  visual object classes (voc) challenge, International journal of computer
  vision 88~(2) (2010) 303--338.

\bibitem{paszke2019pytorch}
A.~Paszke, S.~Gross, F.~Massa, A.~Lerer, J.~Bradbury, G.~Chanan, T.~Killeen,
  Z.~Lin, N.~Gimelshein, L.~Antiga, et~al., Pytorch: An imperative style,
  high-performance deep learning library, in: Advances in neural information
  processing systems, 2019, pp. 8026--8037.

\bibitem{chen2019mmdetection}
K.~Chen, J.~Wang, J.~Pang, Y.~Cao, Y.~Xiong, X.~Li, S.~Sun, W.~Feng, Z.~Liu,
  J.~Xu, et~al., Mmdetection: Open mmlab detection toolbox and benchmark, arXiv
  preprint arXiv:1906.07155 (2019).

\bibitem{berthelot2019mixmatch}
D.~Berthelot, N.~Carlini, I.~Goodfellow, N.~Papernot, A.~Oliver, C.~A. Raffel,
  Mixmatch: A holistic approach to semi-supervised learning, in: Advances in
  Neural Information Processing Systems, 2019, pp. 5049--5059.

\bibitem{huang2021behavior}
S.~Huang, X.~Zeng, S.~Wu, Z.~Yu, M.~Azzam, H.-S. Wong, Behavior regularized
  prototypical networks for semi-supervised few-shot image classification,
  Pattern Recognition 112 (2021) 107765.

\bibitem{lai2019improving}
D.~Lai, W.~Tian, L.~Chen, Improving classification with semi-supervised and
  fine-grained learning, Pattern Recognition 88 (2019) 547--556.

\bibitem{redmon2018yolov3}
J.~Redmon, A.~Farhadi, Yolov3: An incremental improvement, arXiv preprint
  arXiv:1804.02767 (2018).

\end{thebibliography}
\end{small}





% \end{thebibliography}

\end{document}